\pdfoutput=1

\documentclass[11pt]{article}
\usepackage[final]{acl}
\usepackage[T1]{fontenc}    
\usepackage{hyperref}       
\usepackage{url}            
\usepackage{booktabs}       
\usepackage{amsfonts}       
\usepackage{amsmath}        
\usepackage{nicefrac}       
\usepackage{microtype}      
\usepackage[table]{xcolor}         
\usepackage{multirow}
\usepackage{tcolorbox}

\usepackage{subcaption}

\usepackage{graphicx}
\usepackage{longtable}
\usepackage{caption}
\usepackage{adjustbox}
\usepackage{float}      

\usepackage{colortbl}  
\usepackage{array}     
\usepackage{amsmath}   
\usepackage{booktabs}  
\usepackage{makecell}  
\usepackage{wrapfig}   

\usepackage{graphicx}

\usepackage{url}
\usepackage{arydshln}
\usepackage{bm}

\usepackage{times}
\usepackage{latexsym}
\usepackage{enumitem}
\usepackage{array}
\newcommand\ourmethod{\textsc{LsrIF}\xspace}

\usepackage[T1]{fontenc}

\usepackage{microtype}

\usepackage{inconsolata}
\usepackage{makecell}
\usepackage{graphicx}
\usepackage{xspace}
\usepackage{bbm}

\usepackage{algorithm}
\usepackage{algpseudocode}
\usepackage{amsmath}

\title{\ourmethod: Enhancing Logic-Structured Instruction Following \\ of Large Language Models}

\author{
 \textbf{Qingyu Ren\textsuperscript{1}},
 \textbf{Qianyu He\textsuperscript{1}},
 \textbf{Jingwen Chang\textsuperscript{1}}, \textbf{Geng Zhang\textsuperscript{1}}, \textbf{Jiajie Zhu\textsuperscript{1}}, \textbf{Xingzhou Chen\textsuperscript{1}},\\\textbf{Zhuofei Shi\textsuperscript{1}},
 \textbf{Jiaqing Liang\textsuperscript{2}\thanks{\ Corresponding author.}},
 \textbf{Yanghua Xiao\textsuperscript{1}\footnotemark[1]}, \textbf{Han Xia\textsuperscript{3}, Zeye Sun\textsuperscript{3}, Fei Yu\textsuperscript{3}}\
\\
    \textsuperscript{\rm 1}Shanghai Key Laboratory of Data Science, \\College of Computer Science and Artificial Intelligence, Fudan University,\\
    \textsuperscript{\rm 2}School of Data Science, Fudan University,
    \textsuperscript{\rm 3}Ant Group\\
     \{qyren24,qyhe21\}@m.fudan.edu.cn, \{liangjiaqing, shawyh\}@fudan.edu.cn
}

\begin{document}
\maketitle
\begin{abstract}
Instruction following is critical for large language models, yet real-world instructions often involve multiple constraints with logical structures, such as parallel composition, sequential dependencies, and conditional branching. Existing methods typically construct data by simply combining constraints and aggregate rewards by averaging individual constraint scores during training, overlooking logical dependencies and introducing noisy signals. We propose \ourmethod, a  training framework for logic-structured instruction following. \ourmethod constructs data by organizing atomic constraints into parallel, sequential, conditional, and nested structures, and applies structure-aware reward aggregation aligned with their execution semantics: averaging rewards for parallel constraints, decaying later rewards after early failures in sequential structures, and rewarding only active branches in conditional structures. Experiments show that \ourmethod improves instruction following in both in-domain and out-of-domain settings while also benefiting logic reasoning. Further analysis indicates that logic-structured training increases attention to constraint-related tokens and logical connectors, suggesting improved modeling of instruction logic. We will release our data and code for future research.

\end{abstract}

\section{Introduction}
\begin{figure}[t] 
    \centering
            \includegraphics[width=0.45\textwidth]{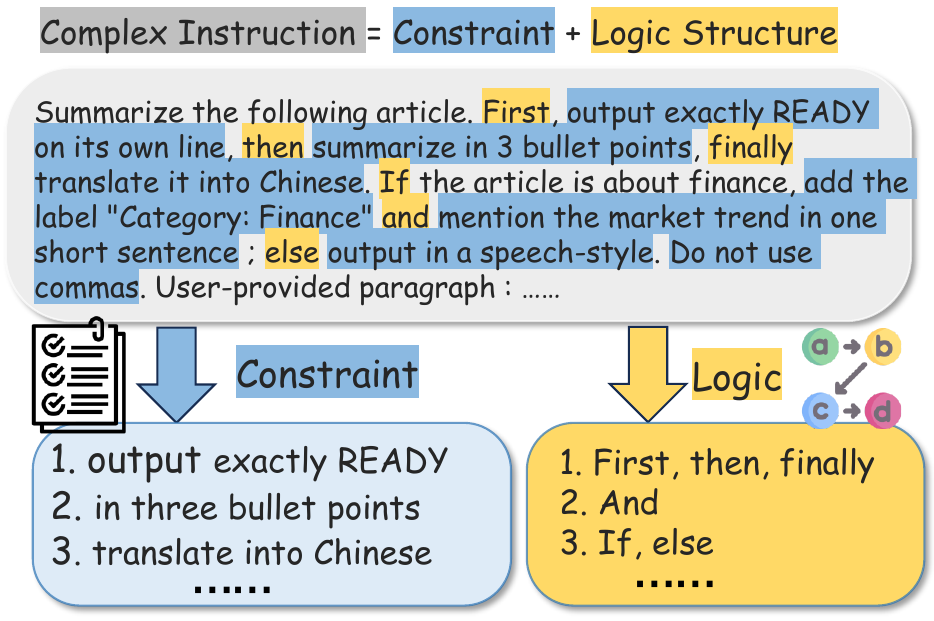}
    \caption{Essentially, the complex instruction is the logical composition of constraints.}
    \label{fig:motivation}
\end{figure}

\begin{figure*}[t] 
    \centering
            \includegraphics[width=\textwidth]{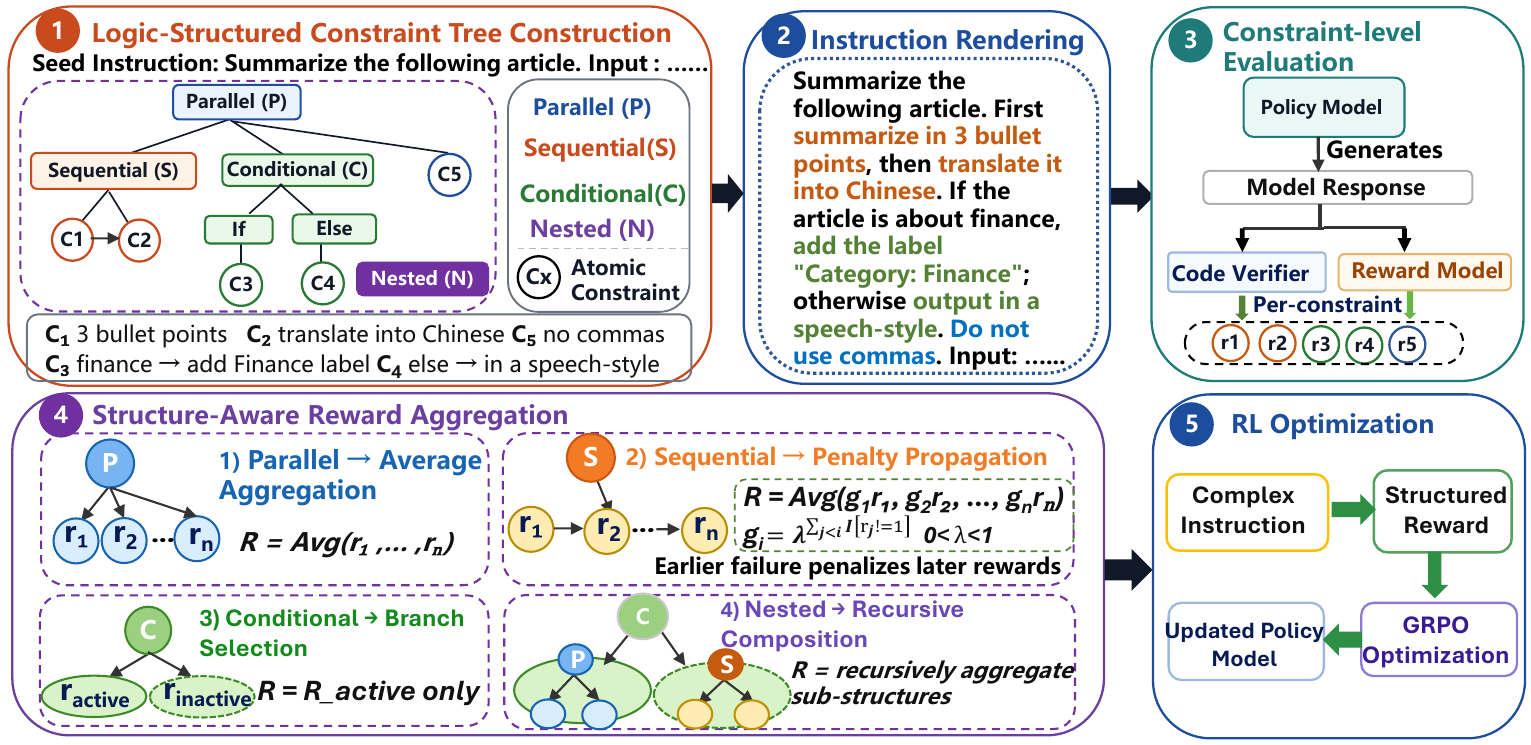}
    \caption{Overview of \ourmethod. \ourmethod first constructs a logic-structured constraint tree from atomic constraints and their logic structures and renders it into a complex instruction. The model response is evaluated at the constraint level, and rewards are aggregated according to tree semantics, including parallel averaging, sequential penalty propagation, conditional branch selection, and recursive nested aggregation. The resulting structured reward is used for policy optimization.}

    \label{fig:framework}
\end{figure*}

Instruction following is a core capability of large language models (LLMs) and is essential to their deployment in real-world applications~\cite{zhang2025recommendation,lu2025developing,ye2025multi,liu2026ministral}. In practice, user instructions are often complex, containing multiple constraints on the model's response~\cite{qi2025agentif,deshpande2025multichallenge,li2026thinking}. Effective instruction following therefore requires models to go beyond producing fluent text: they must accurately understand and satisfy multiple constraints in the instruction~\cite{he2024complex,an2025ultraif,pyatkin2025generalizing}.

In many cases, constraints are not independent but connected through logical structures. Correct instruction following therefore requires not only satisfying individual constraints, but also adhering to the logical relationships among them. Fig.~\ref{fig:motivation} illustrates three  types of logical relationships in complex instructions. Parallel (And) structures require all constraints to be satisfied simultaneously. Sequential (First--Then--Finally) structures impose an execution order, where later constraints depend on the successful completion of earlier ones. Conditional (If--Else) structures introduce branching logic, where the model must first evaluate the condition and then follow the active branch.

Existing approaches for improving instruction following still face clear limitations in handling logic-structured instructions. From the perspective of \textbf{data construction}, most existing methods overlook the logical relationships among constraints. They typically construct training data by directly adding multiple constraints to an instruction, while ignoring more complex structures such as sequential dependencies and conditional branches~\cite{sun2024conifer,huang2025musc}. Although some datasets consider logical structures, they are mainly designed for evaluation rather than training~\cite{wen2024benchmarking,wang2025complexbench}. In terms of \textbf{reward modeling}, existing methods also fail to account for the logical relationships among constraints. The reward for an entire instruction is often computed as the average of the rewards for individual constraints~\cite{qin2025incentivizing}, assuming that all constraints are independent. However, in sequential structures, failure at an early step can make later constraints irrelevant; in conditional structures, constraints in inactive branches should not contribute to the reward. Simple averaging may therefore produce misleading training signals. Finally, existing methods provide limited \textbf{interpretability analysis} of performance improvements. They rarely investigate why these improvements occur~\cite{peng2025verif}. As a result, it remains unclear what factors drive the observed gains, especially in logically structured instruction following.

To address these limitations, we propose \ourmethod, a training framework that explicitly models the logical relationships among constraints in both data construction and reward design, as shown in Fig.~\ref{fig:framework}. 
(1) \textbf{Logic-Structured Data Construction}. 
Given a seed instruction, \ourmethod selects compatible atomic constraints and organizes them with logic structures such as parallel, sequential, and conditional composition. These structures can be recursively combined into a logic-structured constraint tree, enabling the construction of complex instructions with nested structures. 
(2) \textbf{Structure-Aware Reward Aggregation}. 
We design a structure-aware reward aggregation method that follows the execution semantics of the constructed constraint tree. For parallel structures, rewards from all child constraints are averaged. For sequential structures, we introduce a penalty propagation mechanism, where failures in earlier steps penalize the rewards assigned to later constraints. For conditional structures, only the active branch contributes to the final reward. By recursively applying these aggregation rules over the constraint tree, our reward modeling method provides reward signals that better align with the intended logical structure of the instruction.
(3) \textbf{Interpretability for Performance Improvements}. 
We further analyze why our training improves model performance. Our analysis suggests that the observed gains are closely associated with changes in the attention mechanism. At the token level, trained models place more attention on constraint-related tokens and logical connectors. Similar attention changes also appear in general reasoning tasks, suggesting that the learned ability transfers beyond instruction following.

Our contributions are summarized as follows: 
(1) We propose \ourmethod, a training framework that explicitly models logical relationships among constraints for complex instruction following. 
(2) We introduce logic-structured data construction and structure-aware reward aggregation, enabling both synthesized instructions and reward signals to capture parallel, sequential, conditional, and nested constraint structures. 
(3) We provide an interpretability analysis showing that our training changes the model's attention behavior, increasing attention to constraint-related tokens and logical connectors, with evidence of transfer to logic reasoning tasks.

\section{Related Work}

\subsection{Data Construction for Instruction Following}

Existing work constructs datasets with multi-constraint instructions to improve instruction-following capabilities. However, these approaches ignore logic structures among the constraints, which fails to simulate real-world user instructions~\cite{huang2025musc,cheng2024spar}. While some datasets consider logical structures~\cite{wen2024benchmarking,wang2025complexbench}, they are primarily designed for evaluation rather than training. However, we construct a training dataset where constraints show explicit logical structures.

\subsection{Reward Modeling for Instruction Following}

Training paradigms for instruction following have evolved from Supervised Fine-tuning~\cite{sun2024conifer} to Direct Preference Optimization ~\cite{huang2025musc,qi2024constraint} and Reinforcement Learning with Verifiable Rewards (RLVR)~\cite{peng2025verif,qin2025incentivizing}. Existing  RLVR methods aggregate constraint-level rewards through simple averaging~\cite{liu2025recast,wang2026light}. However, this fails when constraint logical structures are not parallel (e.g., sequential or conditional). Unlike these methods, we propose structure-aware reward modeling.

\section{Method}

\subsection{Overview}

Figure~\ref{fig:framework} illustrates the overall framework of \ourmethod.
Given a seed example consisting of an instruction and an input, we first select task-compatible atomic constraints and organize them with one or more logical structures.
The constraints are composed into a constraint tree, which is then rendered together with the seed example into a complex natural-language instruction.
After the policy model generates a response, we evaluate the response on each atomic constraint and aggregate the constraint-level rewards according to the logic structure of the tree.
The resulting structured reward is used for policy optimization.

\subsection{Problem Formulation}

Let $x^{\mathrm{seed}}=(x^{\mathrm{inst}}, x^{\mathrm{input}})$ denote a seed example, where $x^{\mathrm{inst}}$ is the original instruction and $x^{\mathrm{input}}$ is the corresponding input.
Let $x$ denote the constructed complex instruction, and $y$ the response generated by the policy model $\pi_\theta(y\mid x)$.
Each instruction is associated with a constraint tree $\mathcal{T}_x=(\mathcal{V},\mathcal{E},\rho)$, where $\mathcal{V}$ and $\mathcal{E}$ are the node and edge sets, and $\rho$ is the root node.
Leaf nodes represent atomic constraints, while internal nodes represent logical operators.

We consider four logic types: parallel, sequential, conditional, and nested.
For an internal node $v$, its logic type is denoted by
$\tau(v)\in\{\textsc{Par},\textsc{Seq},\textsc{Cond},\textsc{Nest}\}$.
Parallel, sequential, and conditional nodes describe basic logical relations among constraints.
A nested node represents a higher-level composition of multiple logical substructures.
That is, the children of a nested node can themselves be parallel, sequential, conditional, or nested nodes.
The goal is to train the policy model to satisfy not only individual constraints but also their intended logical structures.

\subsection{Logic-Structured Instruction Construction}

To construct training data with explicit logical structures, we use GPT-4.1 to transform seed examples into complex instructions.
For each seed example $x^{\mathrm{seed}}=(x^{\mathrm{inst}}, x^{\mathrm{input}})$, GPT-4.1 first selects a set of compatible atomic constraints and one or more logical structures.
For conditional structures, GPT-4.1 generates a branch condition based on $x^{\mathrm{input}}$. These constraints and structures are then composed into a constraint tree $\mathcal{T}_x$, and the final instruction $x$ is obtained by rendering the original seed instruction, the input, and the tree into a complex natural-language instruction.

Each training example contains the rendered instruction $x$, its constraint tree $\mathcal{T}_x$, the set of atomic constraints $\mathcal{C}_x$, and the corresponding evaluators $\mathcal{Q}_x$.
In a parallel structure, all child constraints should be satisfied simultaneously.
In a sequential structure, constraints should be satisfied in order, and later constraints depend on earlier ones.
In a conditional structure, only the branch activated by the condition is evaluated.

\subsection{Constraint-Level Evaluation}

Given an instruction $x$ and a response $y$, we first evaluate $y$ on each atomic constraint.
For each leaf node $\ell$ in the constraint tree, let $c_\ell$ be its atomic constraint and $q_\ell$ be the corresponding evaluator.
The constraint-level reward is defined as $r_\ell=q_\ell(x,y,c_\ell)$, where $r_\ell\in \{0,1\}$. For hard constraints, such as format, length, or keyword requirements, we use deterministic programmatic verifiers.
For soft constraints, such as style or semantic requirements, we use a trained reward model.
These leaf-level rewards are then aggregated according to the logical structure of the constraint tree.

\subsection{Structure-Aware Reward Aggregation}

One way to compute the final reward is to average all atomic rewards.
However, this ignores the logical structure of the instruction.
For example, inactive branches in a conditional instruction should not affect the reward.
Therefore, we define a recursive aggregation function $\mathcal{A}(v)$ on the constraint tree.
For a leaf node $\ell$, $\mathcal{A}(\ell)=r_\ell$.
For an internal node, the aggregation depends on its logic type.

\paragraph{Parallel structure.}
For a parallel node, all child nodes are equally required.
Given its children $\mathrm{Ch}(v)=(u_1,\ldots,u_m)$, we compute the average reward:
\begin{equation}
    \mathcal{A}(v)
    =
    \frac{1}{m}
    \sum_{j=1}^{m}
    \mathcal{A}(u_j),
    \quad
    \tau(v)=\textsc{Par}.
\end{equation}

\paragraph{Sequential structure.}
For a sequential node, child nodes should be satisfied in order.
Let $a_j=\mathcal{A}(u_j)$ be the reward of the $j$-th child.
We use an indicator $z_j$ to denote whether the $j$-th child is fully satisfied.
For each constraint, $z_j=1$ if $a_j=1$ and $z_j=0$ otherwise. To penalize later constraints when earlier constraints are not satisfied, we assign each child a penalty $g_j=\lambda^{\sum_{k<j}(1-z_k)}$, where $\lambda\in(0,1)$ controls the penalty strength.
The sequential reward is:
\begin{equation}
    \mathcal{A}(v)
    =
    \frac{1}{m}
    \sum_{j=1}^{m}
    g_j a_j,
    \quad
    \tau(v)=\textsc{Seq}.
\end{equation}
A smaller $\lambda$ gives stronger penalties to later constraints after earlier failures. We present sensitivity analysis on the penalty factor $\lambda$ in Appendix~\ref{appx:sen}.

\paragraph{Conditional structure.}
For a conditional node $v$, let $s_v$ denote the branch condition generated during instruction construction, and let $u_{v,\mathrm{true}}$ and $u_{v,\mathrm{false}}$ denote its true and false branches, respectively.
During reward aggregation, GPT-4.1 evaluates whether the condition $s_v$ holds based on $x^{\mathrm{input}}$:
$
b_v(x^{\mathrm{input}};s_v)\in\{\mathrm{true},\mathrm{false}\}.
$
The activated branch is defined as
\[
u_{v,\mathrm{act}}
=
\begin{cases}
u_{v,\mathrm{true}}, & b_v(x;s_v)=\mathrm{true},\\
u_{v,\mathrm{false}}, & b_v(x;s_v)=\mathrm{false}.
\end{cases}
\]
The conditional reward is then computed as
\begin{equation}
    \mathcal{A}(v)
    =
    \mathcal{A}(u_{v,\mathrm{act}}),
    \quad
    \tau(v)=\textsc{Cond}.
\end{equation}

\paragraph{Nested structures.}
A nested node represents a composition of logical substructures, where each child may itself be a parallel, sequential, conditional, or nested node.
We handle nesting by recursively applying each child's own aggregation rule. This enables a single instruction to combine multiple logical structures, whose rewards are evaluated according to their own semantics before being passed to the parent node.

\paragraph{Final reward.}
The aggregation function is applied recursively from the leaves to the root.
The final structured reward is the value at the root node:
$R_{\mathrm{final}}(x,y)=\mathcal{A}(\rho)$.

\subsection{RL Optimization}

We use the structured reward $R_{\mathrm{final}}$ to optimize the policy model with GRPO~\cite{shao2024deepseekmath}.
For each instruction $x$, we sample a group of $G$ responses $\{y_i\}_{i=1}^{G}$ from the current policy, where $y_i\sim\pi_\theta(\cdot\mid x)$.
For each response, we compute its structured reward as $R_i=R_{\mathrm{final}}(x,y_i)$.
Following GRPO, rewards are normalized within the same group to obtain relative advantages $\hat{A}_i=(R_i-\mu_R)/(\sigma_R+\delta)$, where $\mu_R$ and $\sigma_R$ are the mean and standard deviation of $\{R_i\}_{i=1}^{G}$, and $\delta$ is a small constant for numerical stability.

The normalized advantage $\hat{A}_i$ is then used as the training signal in the GRPO clipped update.
Responses with higher structured rewards receive larger positive advantages and are therefore encouraged, while lower-reward responses are discouraged.
We also regularize the updated policy toward a reference policy $\pi_{\mathrm{ref}}$ to maintain training stability.
In this way, the model is encouraged to generate responses that follow the execution semantics of atomic constraints. We present the pseudocode of our method in Algorithm~\ref{alg:logic_structured_training}.

\label{sec:pse}

\definecolor{lightblue}{RGB}{173, 216, 230}

\definecolor{lightpink}{RGB}{255, 220, 180}

\definecolor{lightgray}{RGB}{211, 211, 211}

\definecolor{lightgreen}{RGB}{144, 238, 144}

\definecolor{lightyellow}{RGB}{255, 255, 200}

\definecolor{lightred}{RGB}{255, 182, 193}

\definecolor{lightpurple}{RGB}{216, 191, 216}

\definecolor{tableblue}{RGB}{123, 166, 180}

\definecolor{darkpink}{RGB}{255, 165, 100}

\definecolor{darkgray}{RGB}{169, 169, 169}

\definecolor{darkgreen}{RGB}{94, 188, 94}

\definecolor{darkyellow}{RGB}{218, 165, 32}

\definecolor{darkred}{RGB}{205, 92, 92}

\definecolor{darkpurple}{RGB}{166, 141, 166}

\begin{table*}[t]

\centering

{\fontsize{8.8pt}{8.6pt}\selectfont

\setlength{\tabcolsep}{3.3pt}
\renewcommand{\arraystretch}{1.2}
\begin{tabular}{llcccccc}

\toprule

\multirow{3}{*}{\textbf{Models}} &

\multirow{3}{*}{\textbf{Method}} &

\multicolumn{3}{c}{\textbf{In-Domain}} &

\multicolumn{3}{c}{\textbf{Out-of-Domain}} \\

\cmidrule(lr){3-5} \cmidrule(lr){6-8}

& & \textbf{IFEval} & \textbf{CFBench} & \textbf{FollowBench}

& \textbf{ComplexBench} & \textbf{Collie} & \textbf{AgentIF} \\

\cmidrule(lr){3-5} \cmidrule(lr){6-8}

& & \textbf{Pr.(L)} & \textbf{ISR} & \textbf{HSR}

& \textbf{Overall} & \textbf{Avg.} & \textbf{CSR} \\

\midrule

GPT-4o

& \cellcolor{darkpink!25}Baseline

& \cellcolor{lightpink!25}84.8 & \cellcolor{lightpink!25}65.3 & \cellcolor{lightpink!25}70.4 & \cellcolor{lightpink!25}71.6 & \cellcolor{lightpink!25}49.8 & \cellcolor{lightpink!25}58.5 \\

QwQ-32B

& \cellcolor{darkpink!25}Baseline

& \cellcolor{lightpink!25}83.9 & \cellcolor{lightpink!25}68.0 & \cellcolor{lightpink!25}62.2 & \cellcolor{lightpink!25}73.3 & \cellcolor{lightpink!25}52.4 & \cellcolor{lightpink!25}58.1 \\

Self-Supervised-7B

& \cellcolor{darkpink!25}Baseline

& \cellcolor{lightpink!25}78.9 & \cellcolor{lightpink!25}52.0 & \cellcolor{lightpink!25}57.5 & \cellcolor{lightpink!25}68.7 & \cellcolor{lightpink!25}38.0 & \cellcolor{lightpink!25}56.7 \\

VERIF-8B

& \cellcolor{darkpink!25}Baseline

& \cellcolor{lightpink!25}87.1 & \cellcolor{lightpink!25}41.0 & \cellcolor{lightpink!25}56.9 & \cellcolor{lightpink!25}54.7 & \cellcolor{lightpink!25}28.3 & \cellcolor{lightpink!25}56.6 \\

RAIF-7B& \cellcolor{darkpink!25}Baseline

& \cellcolor{lightpink!25}74.1 & \cellcolor{lightpink!25}43.0 & \cellcolor{lightpink!25}56.2 & \cellcolor{lightpink!25}68.7 & \cellcolor{lightpink!25}20.2 & \cellcolor{lightpink!25}51.9 \\

SPAR-8B-DPO

& \cellcolor{darkpink!25}Baseline

& \cellcolor{lightpink!25}82.4 & \cellcolor{lightpink!25}37.0 & \cellcolor{lightpink!25}56.1 & \cellcolor{lightpink!25}63.8 & \cellcolor{lightpink!25}27.7 & \cellcolor{lightpink!25}53.6 \\

Crab-7B-DPO

& \cellcolor{darkpink!25}Baseline

& \cellcolor{lightpink!25}57.7 & \cellcolor{lightpink!25}25.0 & \cellcolor{lightpink!25}49.4 & \cellcolor{lightpink!25}59.0 & \cellcolor{lightpink!25}19.6 & \cellcolor{lightpink!25}47.2 \\

Conifer-7B-DPO

& \cellcolor{darkpink!25}Baseline

& \cellcolor{lightpink!25}52.3 & \cellcolor{lightpink!25}25.0 & \cellcolor{lightpink!25}50.0 & \cellcolor{lightpink!25}48.1 & \cellcolor{lightpink!25}17.8 & \cellcolor{lightpink!25}44.3 \\

\midrule

\multirow{3}{*}{Qwen2.5-1.5B-Instruct}

& \cellcolor{darkpink!25}Base

& \cellcolor{lightpink!25}43.6 & \cellcolor{lightpink!25}22.0 & \cellcolor{lightpink!25}34.6 & \cellcolor{lightpink!25}45.9 & \cellcolor{lightpink!25}13.0 & \cellcolor{lightpink!25}42.8 \\

& \cellcolor{darkpink!25}SFT

& \cellcolor{lightpink!25}64.0 & \cellcolor{lightpink!25}24.0& \cellcolor{lightpink!25}37.4 & \cellcolor{lightpink!25}49.8 & \cellcolor{lightpink!25}16.1 & \cellcolor{lightpink!25}46.4 \\

& \cellcolor{darkpink!25}\ourmethod

& \cellcolor{lightpink!25}68.8 \textcolor{red}{(+25.2)} & \cellcolor{lightpink!25}28.0 \textcolor{red}{(+6.0)} & \cellcolor{lightpink!25}38.9 \textcolor{red}{(+4.3)} & \cellcolor{lightpink!25}52.4 \textcolor{red}{(+6.5)} & \cellcolor{lightpink!25}19.3 \textcolor{red}{(+6.3)} & \cellcolor{lightpink!25}51.5 \textcolor{red}{(+8.7)} \\

\midrule

\multirow{3}{*}{Qwen2.5-7B-Instruct}

& \cellcolor{tableblue!25}Base

& \cellcolor{lightblue!25}73.9 & \cellcolor{lightblue!25}47.0 & \cellcolor{lightblue!25}55.1 & \cellcolor{lightblue!25}66.1 & \cellcolor{lightblue!25}36.3 & \cellcolor{lightblue!25}54.2 \\

& \cellcolor{tableblue!25}SFT

& \cellcolor{lightblue!25}75.2 & \cellcolor{lightblue!25}43.0 & \cellcolor{lightblue!25}55.7 & \cellcolor{lightblue!25}68.5 & \cellcolor{lightblue!25}30.5 & \cellcolor{lightblue!25}55.5 \\

& \cellcolor{tableblue!25}\ourmethod

& \cellcolor{lightblue!25}79.7 \textcolor{red}{(+5.8)} & \cellcolor{lightblue!25}54.0 \textcolor{red}{(+7.0)} & \cellcolor{lightblue!25}57.5 \textcolor{red}{(+2.4)} & \cellcolor{lightblue!25}70.0 \textcolor{red}{(+3.9)} & \cellcolor{lightblue!25}37.3 \textcolor{red}{(+1.0)} & \cellcolor{lightblue!25}56.5 \textcolor{red}{(+2.3)} \\

\midrule

\multirow{3}{*}{Distill-Qwen-7B}

& \cellcolor{darkgray!25}Base

& \cellcolor{lightgray!25}61.7 & \cellcolor{lightgray!25}36.0 & \cellcolor{lightgray!25}41.7 & \cellcolor{lightgray!25}55.2 & \cellcolor{lightgray!25}25.2 & \cellcolor{lightgray!25}47.2 \\

& \cellcolor{darkgray!25}SFT

& \cellcolor{lightgray!25}65.1 & \cellcolor{lightgray!25}40.0 & \cellcolor{lightgray!25}43.1 & \cellcolor{lightgray!25}55.8 & \cellcolor{lightgray!25}28.3 & \cellcolor{lightgray!25}44.2 \\

& \cellcolor{darkgray!25}\ourmethod

& \cellcolor{lightgray!25}71.5 \textcolor{red}{(+9.8)} & \cellcolor{lightgray!25}47.0 \textcolor{red}{(+11.0)} & \cellcolor{lightgray!25}44.0 \textcolor{red}{(+2.3)} & \cellcolor{lightgray!25}61.1 \textcolor{red}{(+5.9)} & \cellcolor{lightgray!25}30.0 \textcolor{red}{(+4.8)} & \cellcolor{lightgray!25}46.7 \textcolor{gray}{(-0.5)} \\

\midrule

\multirow{3}{*}{Llama-3.1-8B-Instruct}

& \cellcolor{darkgreen!25}Base

& \cellcolor{lightgreen!25}73.8 & \cellcolor{lightgreen!25}34.0 & \cellcolor{lightgreen!25}53.8 & \cellcolor{lightgreen!25}63.6 & \cellcolor{lightgreen!25}46.5 & \cellcolor{lightgreen!25}53.4 \\

& \cellcolor{darkgreen!25}SFT

& \cellcolor{lightgreen!25}77.4 & \cellcolor{lightgreen!25}36.0 & \cellcolor{lightgreen!25}52.2 & \cellcolor{lightgreen!25}61.1 & \cellcolor{lightgreen!25}34.5 & \cellcolor{lightgreen!25}55.2 \\

& \cellcolor{darkgreen!25}\ourmethod

& \cellcolor{lightgreen!25}81.5 \textcolor{red}{(+7.7)} & \cellcolor{lightgreen!25}40.0 \textcolor{red}{(+6.0)} & \cellcolor{lightgreen!25}58.4 \textcolor{red}{(+4.6)} & \cellcolor{lightgreen!25}63.9 \textcolor{red}{(+0.3)} & \cellcolor{lightgreen!25}47.6 \textcolor{red}{(+1.1)} & \cellcolor{lightgreen!25}57.8 \textcolor{red}{(+4.4)} \\

\midrule

\multirow{3}{*}{Qwen3-8B}

& \cellcolor{darkpurple!25}Base

& \cellcolor{lightpurple!25}87.8 & \cellcolor{lightpurple!25}66.0 & \cellcolor{lightpurple!25}56.4 & \cellcolor{lightpurple!25}78.5 & \cellcolor{lightpurple!25}45.5 & \cellcolor{lightpurple!25}64.4 \\

& \cellcolor{darkpurple!25}SFT

& \cellcolor{lightpurple!25}88.3 & \cellcolor{lightpurple!25}66.0 & \cellcolor{lightpurple!25}56.7 & \cellcolor{lightpurple!25}78.3 & \cellcolor{lightpurple!25}45.7 & \cellcolor{lightpurple!25}63.9 \\

& \cellcolor{darkpurple!25}\ourmethod

& \cellcolor{lightpurple!25}90.2 \textcolor{red}{(+2.4)} & \cellcolor{lightpurple!25}68.0 \textcolor{red}{(+2.0)} & \cellcolor{lightpurple!25}58.1 \textcolor{red}{(+1.7)} & \cellcolor{lightpurple!25}79.2 \textcolor{red}{(+0.7)} & \cellcolor{lightpurple!25}48.1 \textcolor{red}{(+2.6)} & \cellcolor{lightpurple!25}65.0 \textcolor{red}{(+0.6)} \\
\midrule
\multirow{3}{*}{Distill-Qwen-14B}

& \cellcolor{darkyellow!25}Base

& \cellcolor{lightyellow!25}74.9 & \cellcolor{lightyellow!25}55.0 & \cellcolor{lightyellow!25}51.2 & \cellcolor{lightyellow!25}72.7 & \cellcolor{lightyellow!25}34.4 & \cellcolor{lightyellow!25}54.5 \\

& \cellcolor{darkyellow!25}SFT

& \cellcolor{lightyellow!25}79.3 & \cellcolor{lightyellow!25}56.0 & \cellcolor{lightyellow!25}56.8 & \cellcolor{lightyellow!25}70.5 & \cellcolor{lightyellow!25}36.1 & \cellcolor{lightyellow!25}59.2 \\

& \cellcolor{darkyellow!25}\ourmethod

& \cellcolor{lightyellow!25}82.1 \textcolor{red}{(+7.2)} & \cellcolor{lightyellow!25}60.0 \textcolor{red}{(+5.0)} & \cellcolor{lightyellow!25}58.2 \textcolor{red}{(+7.0)} & \cellcolor{lightyellow!25}75.5 \textcolor{red}{(+2.8)} & \cellcolor{lightyellow!25}38.8 \textcolor{red}{(+4.4)} & \cellcolor{lightyellow!25}61.7 \textcolor{red}{(+7.2)} \\

\bottomrule

\end{tabular}}

\caption{Model performance on in-domain and out-of-domain instruction following benchmarks.}

\label{tab:api_overall}


\end{table*}

\section{Experiment}

\subsection{Experimental Setup}

\paragraph{\textbf{Models.}} We conduct experiments on models of  scales from 1.5B to 14B to evaluate the effectiveness of our method across different architectures and parameter scales. We compare against both strong general-purpose models and specialized instruction-following optimized models. Details are provided in Appendix~\ref{appx:bl}.

\paragraph{\textbf{Training Methods.}} We compare three training methods: \textbf{Base} uses the original model directly without any additional training; \textbf{SFT} fine-tunes the model on the dataset generated by the strong models using supervised fine-tuning; \textbf{\ourmethod} is our reinforcement learning training method. For the reward model, we fine-tune Qwen2.5-7B-Instruct using the method and data from \citealp{ren2025instructions}. Details are provided in Appendix~\ref{appx:tds}.

\paragraph{\textbf{Evaluation Benchmarks.}} We evaluate models on both in-domain and out-of-domain instruction following benchmarks. \textbf{In-domain benchmarks} include IFEval~\cite{zhou2023instruction}, CFBench~\cite{zhang2024cfbench}, and FollowBench~\cite{jiang2023followbench}. \textbf{Out-of-domain benchmarks} include ComplexBench~\cite{wen2024benchmarking}, Collie~\cite{yao2023collie}, and AgentIF~\cite{qi2025agentif}. Details are provided in Appendix~\ref{appx:bec}.

\paragraph{\textbf{Training Data.}}
Our dataset contains 23 types of hard constraints and 21 types of soft constraints. Each seed instruction can be augmented with one to five atomic constraints, which are organized using one or more logical structures. Finally, the dataset contains 17,510 parallel logical structures, 10,435 sequential logical structures, and 10,574 conditional logical structures. Details are provided in Appendix~\ref{appx:ctd}.

\begin{table*}[t]
\centering
\setlength{\tabcolsep}{2pt}
\resizebox{\textwidth}{!}{
\begin{tabular}{lccccccccc}
\toprule
\multirow{2}{*}{\textbf{Model}} 
& \multicolumn{4}{c}{\textbf{Logic Reasoning (Enigmata)}} 
& \multicolumn{5}{c}{\textbf{General Capabilities}} \\
\cmidrule(lr){2-5} \cmidrule(lr){6-10}
& \textbf{Logic} & \textbf{Arithmetic} & \textbf{Graph} & \textbf{Search} 
& \textbf{AIME2024} & \textbf{AIME2025} & \textbf{GPQA-Diamond} & \textbf{MT-Bench} & \textbf{AlpacaEval2.0} \\
\midrule
Distill-Qwen-7B  
& \cellcolor{lightblue!25}10.9 & \cellcolor{lightblue!25}3.7 & \cellcolor{lightblue!25}11.1 & \cellcolor{lightblue!25}4.4
& \cellcolor{lightblue!25}53.4 & \cellcolor{lightblue!25}38.7 & \cellcolor{lightblue!25}49.1 & \cellcolor{lightblue!25}5.9 & \cellcolor{lightblue!25}5.0 \\
Distill-Qwen-7B-\ourmethod  
& \cellcolor{lightblue!25}\textbf{13.6} & \cellcolor{lightblue!25}\textbf{14.3} \textcolor{red}{\scriptsize(+10.6)}& \cellcolor{lightblue!25}\textbf{17.5} & \cellcolor{lightblue!25}\textbf{4.6}
& \cellcolor{lightblue!25}\textbf{55.1} & \cellcolor{lightblue!25}\textbf{41.2} & \cellcolor{lightblue!25}\textbf{52.5} & \cellcolor{lightblue!25}\textbf{6.3} & \cellcolor{lightblue!25}\textbf{5.8} \\
\midrule
Distill-Qwen-14B 
& \cellcolor{lightpurple!25}44.7 & \cellcolor{lightpurple!25}21.0 & \cellcolor{lightpurple!25}31.1 & \cellcolor{lightpurple!25}10.5
& \cellcolor{lightpurple!25}69.3 & \cellcolor{lightpurple!25}49.0 & \cellcolor{lightpurple!25}58.6 & \cellcolor{lightpurple!25}6.6 & \cellcolor{lightpurple!25}26.7 \\
Distill-Qwen-14B-\ourmethod 
& \cellcolor{lightpurple!25}\textbf{48.4} & \cellcolor{lightpurple!25}\textbf{39.0} \textcolor{red}{\scriptsize(+18.0)} & \cellcolor{lightpurple!25}\textbf{33.3} & \cellcolor{lightpurple!25}\textbf{14.1}
& \cellcolor{lightpurple!25}\textbf{70.2} & \cellcolor{lightpurple!25}\textbf{49.6} & \cellcolor{lightpurple!25}\textbf{60.1} & \cellcolor{lightpurple!25}\textbf{7.0} & \cellcolor{lightpurple!25}\textbf{30.3} \\
\bottomrule
\end{tabular}
}
\caption{
  Model performance on logic reasoning (Enigmata) and general capabilities benchmarks. We evaluate AIME using Avg@30 method. \textbf{Bolded} value indicates the best result for each model on the benchmark.
}
\label{tab:resaon_results}
\end{table*}

\subsection{Main Results}
\paragraph{\textbf{Instruction Following Performance.}}
As shown in Tab.~\ref{tab:api_overall}, \ourmethod improves overall instruction-following performance across different model families and scales. Compared with Base, \ourmethod achieves clear gains on in-domain benchmarks, improving IFEval by $+2.4$ to $+25.2$, CFBench by $+2.0$ to $+11.0$, and FollowBench by $+1.7$ to $+7.0$. The improvements are particularly pronounced for smaller or weaker models: Qwen2.5-1.5B-Instruct improves from $43.6$ to $68.8$ on IFEval and from $42.8$ to $51.5$ on AgentIF, while Distill-Qwen-7B gains $+9.8$ on IFEval and $+11.0$ on CFBench. \ourmethod also brings consistent gains to stronger models, with Qwen3-8B reaching $90.2$ on IFEval and $68.0$ on CFBench.

On out-of-domain benchmarks, \ourmethod also demonstrates strong generalization. It improves ComplexBench by up to $+6.5$, Collie by up to $+6.3$, and AgentIF by up to $+8.7$ over Base. Compared with SFT, \ourmethod achieves higher scores across all evaluated model--benchmark combinations, indicating that it provides more robust instruction-following improvements than standard supervised fine-tuning. Notably, \ourmethod enables open-source models to match or surpass strong baseline models on several benchmarks. For example, Qwen3-8B with \ourmethod achieves $90.2$ on IFEval, outperforming GPT-4o, QwQ-32B, and VERIF-8B, and reaches $79.2$ on ComplexBench and $65.0$ on AgentIF, exceeding all listed baselines on these two out-of-domain benchmarks.

\begin{table}[t]
\resizebox{\columnwidth}{!}{
\fontsize{6.8pt}{8.6pt}\selectfont
\setlength{\tabcolsep}{4pt}
\begin{tabular}{lcccc}
\toprule
\multirow{2}{*}{\textbf{Config}} & \multicolumn{4}{c}{\textbf{Performance}} \\
      & \textbf{IFEval} & \textbf{CFBench} & \textbf{Enigmata-Arith.} \\
\hline
Distill-Qwen-7B
& \cellcolor{lightblue!25}61.7
& \cellcolor{lightblue!25}36.0
& \cellcolor{lightblue!25}3.7
 \\

Distill-Qwen-7B-\ourmethod
& \cellcolor{lightblue!25}\textbf{71.5}
& \cellcolor{lightblue!25}\textbf{47.0}
& \cellcolor{lightblue!25}\textbf{14.3}
\\
\hline
w/o Logic-Structured Data
& \cellcolor{lightpink!25}69.1
& \cellcolor{lightpink!25}45.0
& \cellcolor{lightpink!25}7.7
 \\
w/o Structure-Aware Reward
& \cellcolor{lightpurple!25}67.6
& \cellcolor{lightpurple!25}42.0
& \cellcolor{lightpurple!25}10.7
 \\

w/o Penalty Propagation
& \cellcolor{lightred!25}68.7
& \cellcolor{lightred!25}44.0
& \cellcolor{lightred!25}12.3
 \\

w/o Branch Selection
& \cellcolor{lightgreen!25}67.9
& \cellcolor{lightgreen!25}44.0
& \cellcolor{lightgreen!25}11.3
 \\
\bottomrule
\end{tabular}
}
\caption{Ablation study results.}
\label{tab:abla}
\end{table}

\begin{figure}[!t]
    \centering
    \begin{subfigure}[!t]{\linewidth}
        \centering
    \includegraphics[width=0.8\linewidth]{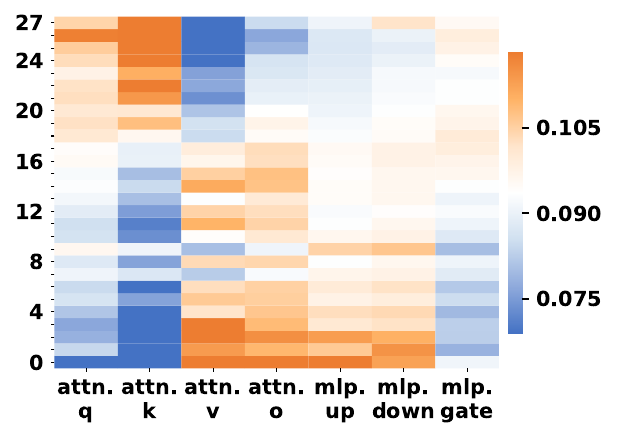}
        \label{fig:param-qwen-7b1}
    \end{subfigure}

    \caption{Parameter change rates of Qwen2.5-7B-Instruct after training. Darker orange colors indicate larger parameter changes.
    }
    \label{fig:parameter}

\end{figure}

\begin{figure*}[t] 
    \centering
    \begin{subfigure}[b]{\textwidth}
        \centering
        \includegraphics[width=\textwidth]{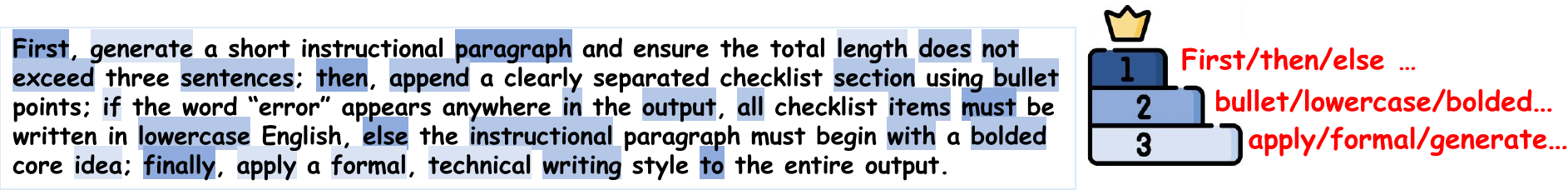}
        \caption{Instruction Following -- More token saliency on constraints and their underlying logic}
        \label{fig:before_inst}
    \end{subfigure}
    \hfill
    \begin{subfigure}[b]{\textwidth}
        \centering
        \includegraphics[width=\textwidth]{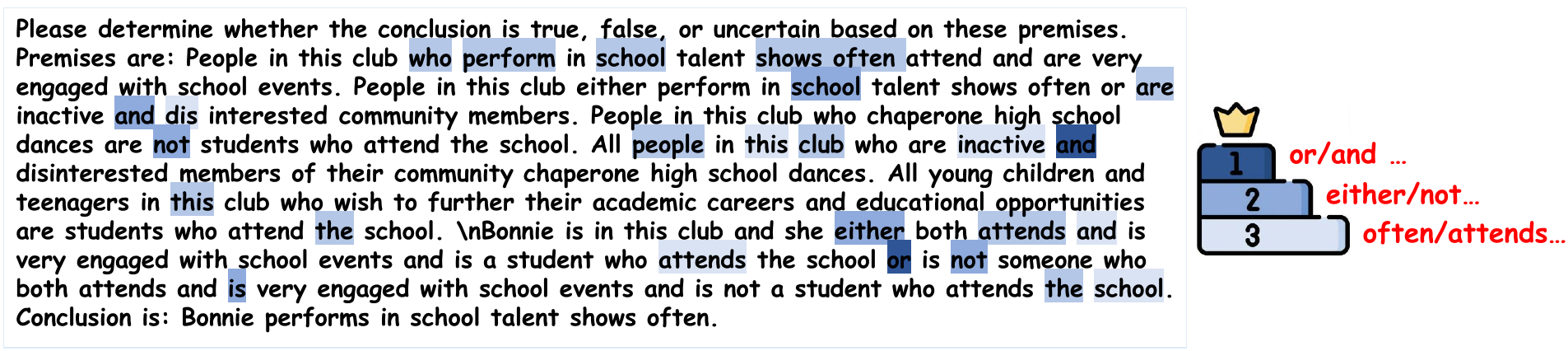}
        \caption{Logic Reasoning -- More token saliency on logical connectors}
        \label{fig:after_inst}
    \end{subfigure}
    
    
    
    \caption{Comparison of gradient-based token saliency changes in Qwen2.5-7B-Instruct before and after training on instruction-following and logical-reasoning tasks. Darker colors indicate larger increases in token saliency.
}
    \label{fig:case11}
\end{figure*}

\paragraph{\textbf{Logical Reasoning Performance.}}

We further evaluate logical reasoning using Enigmata~\cite{chen2025enigmata}, which includes 36 automatically generated and rule-verifiable tasks across seven categories. Following the benchmark setting, we report four representative subcategories: \textbf{Logic}, \textbf{Arithmetic}, \textbf{Graph}, and \textbf{Search}.

As shown in Tab.~\ref{tab:resaon_results}, \ourmethod  improves logical reasoning across all subcategories. The gains are particularly large on Arithmetic, where Distill-Qwen-7B improves from $3.7$ to $14.3$ $(+10.6)$ and Distill-Qwen-14B from $21.0$ to $39.0$ $(+18.0)$. \ourmethod also improves Logic and Graph for both models, as well as Search for Distill-Qwen-14B, indicating stronger structure-sensitive reasoning, especially for numerical constraints.

Beyond reasoning, \ourmethod also improves general capabilities across mathematics, science, and instruction-following benchmarks. For Distill-Qwen-7B, it improves AIME2024, AIME2025, GPQA-Diamond, MT-Bench, and AlpacaEval2.0. Similar consistent gains are observed for Distill-Qwen-14B, including an improvement on AlpacaEval2.0 from $26.7$ to $30.3$. These results suggest that \ourmethod strengthens logical reasoning while preserving and improving general model capabilities.





\subsection{Ablation Studies}

As shown in Tab.~\ref{tab:abla}, all ablation variants underperform the full \ourmethod, demonstrating the importance of modeling logical structures in both data construction and reward design. For data construction, w/o Logic-Structured Data keeps the same training data scale but removes logical organization; instead, each instruction is constructed by directly appending five atomic constraints, and rewards are averaged over all constraints. This causes drops of $2.4$ on IFEval, $2.0$ on CFBench, and $6.6$ on Enigmata-Arithmetic, showing logic-structured data helps the model capture constraint dependencies and improve logic reasoning.

For reward aggregation, w/o Structure-Aware Reward keeps the logic-structured data but averages all atomic rewards, leading to drops of $3.9$, $5.0$, and $3.6$ on the three benchmarks.
Removing penalty propagation further hurts sequential modeling, with drops of $2.8$, $3.0$, and $2.0$.
Removing branch selection also reduces performance by $3.6$, $3.0$, and $3.0$, confirming the need to evaluate only active conditional branches. Overall, the full \ourmethod achieves the best results, validating the necessity of both logic-structured data construction and structure-aware reward aggregation.

\begin{figure*}[t] 
    \centering
            \includegraphics[width=0.95\textwidth]{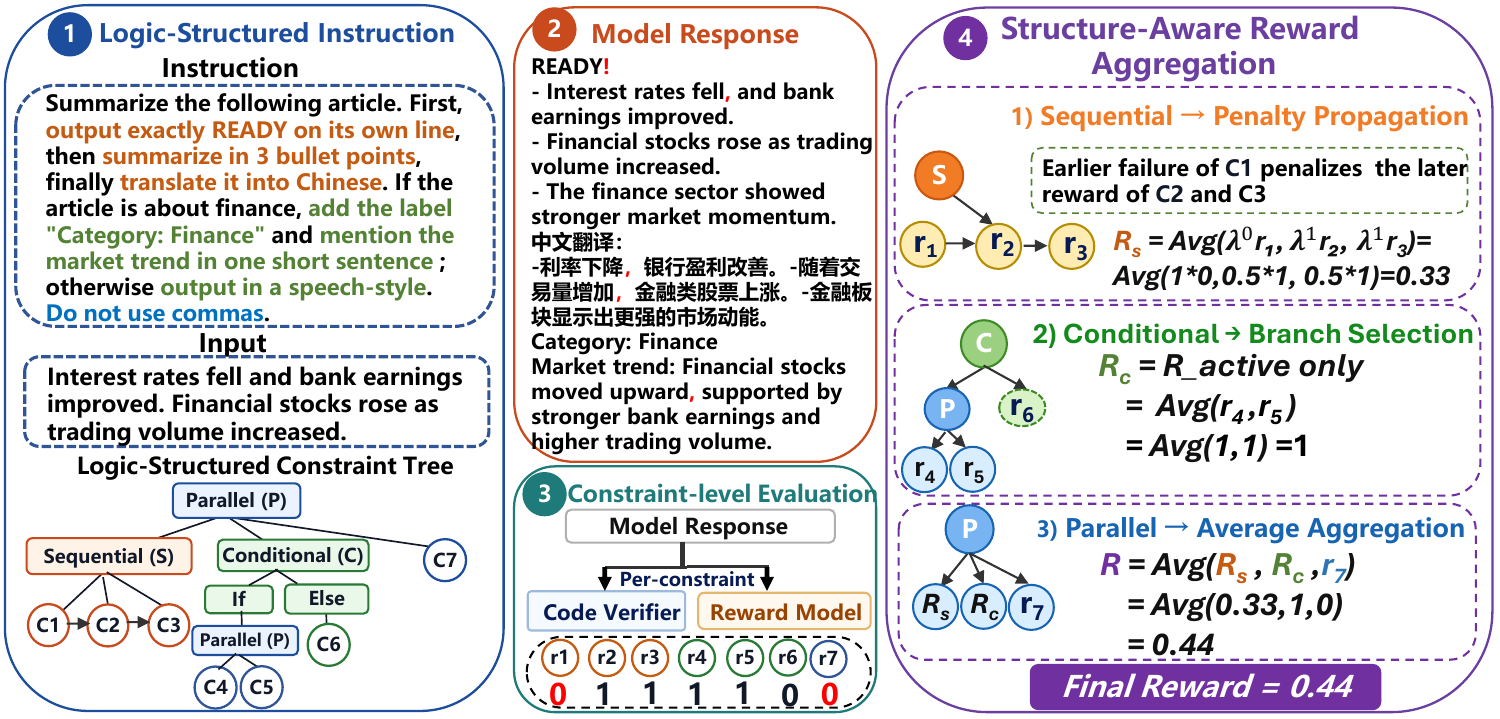}
    \caption{Case study of structure-aware reward aggregation. The instruction is decomposed into sequential, conditional, and parallel constraint structures. \ourmethod first evaluates each atomic constraint, then applies penalty propagation for sequential constraints, active-branch selection for conditional constraints, and average aggregation for parallel constraints to obtain the final reward.}

    \label{fig:case}
\end{figure*}

\subsection{Interpretability Analysis}

\subsubsection{Parameter Change Patterns}

Fig.~\ref{fig:parameter} shows the  parameter changes after training, measured by the normalized Frobenius norm:
$
\Delta = \frac{\|W_{\text{after}} - W_{\text{before}}\|_F}{\|W_{\text{before}}\|_F} \times 100\%,
$
where $W_{\text{before}}$ and $W_{\text{after}}$ denote the parameters before and after training.

The parameter changes exhibit a layer- and module-dependent pattern. In lower layers, relatively larger updates are mainly observed in the attention value/output projections and MLP up/down projections, suggesting that \ourmethod affects low-level feature transformation and information propagation. In upper layers, the attention key projection shows more pronounced changes, while the attention value projection becomes comparatively more stable. This suggests that higher layers may be more involved in adapting token matching and attention routing. Overall, attention-related modules show more structured and heterogeneous changes across depth than MLP modules. In particular, the query/key/value/output projections are updated differently across layers, rather than changing uniformly. These results suggest that \ourmethod induces layer-specific adaptations in attention modules, which may help the model better capture logical dependencies during instruction following. Additional examples are provided in Appendix~\ref{appx:full}.

\subsubsection{Token-Level Information Flow Analysis}

We further analyze token-level information flow using gradient-based saliency. For token $x_i$ with embedding $E_i$, its attribution score is defined as
\begin{equation}
S_i = \left| \sum_{d=1}^{D} \frac{\partial L}{\partial E_{i,d}} E_{i,d} \right|,
\end{equation}
where the sequence-level loss is the negative log-likelihood:
\begin{equation}
L(x,y) = - \sum_{t=1}^{|y|} \log P(y_t \mid y_{<t}, x).
\end{equation}
We measure the change in token importance as
$
\Delta S_i = S_i^{\text{after}} - S_i^{\text{before}}.
$

As shown in Fig.~\ref{fig:case11}, \ourmethod leads to more concentrated saliency patterns in the qualitative examples. For instruction-following tasks, larger increases appear on logical connectors such as ``First'', ``then'', and ``else'', as well as constraint tokens such as ``bullet'', ``lowercase'', and ``bolded''. For logical reasoning tasks, logical operators such as ``or'' and ``and'' receive stronger increases, followed by choice or negation terms such as ``either'' and ``not''. This pattern suggests that \ourmethod encourages the model to assign more importance to tokens encoding logical and constraint structures. Together with the observed changes in attention query and key projections, these results are consistent with the view that \ourmethod improves instruction following by adapting attention mechanisms to better capture constraint relationships. Additional examples are provided in Appendix~\ref{appx:full1}.

\subsection{Case Study}

Fig.~\ref{fig:case} illustrates how \ourmethod evaluates a response under a logic-structured instruction. The instruction is decomposed into a constraint tree with sequential, conditional, and parallel structures, and each leaf constraint is first evaluated at the constraint level. Although the response satisfies several local constraints, such as generating bullet points, translating the summary into Chinese, and identifying the finance category, it fails to output exactly ``READY'' as the first step and also violates the comma-related constraint.

For the sequential structure, \ourmethod propagates the penalty from the failed prerequisite constraint to later dependent constraints, producing $R_s=\mathrm{Avg}(1\times0,0.5\times1,0.5\times1)=0.33$. For the conditional structure, since the article is about finance, only the active finance branch is evaluated, while the inactive branch is ignored, yielding $R_c=\mathrm{Avg}(r_4,r_5)=1$. Finally, the parallel root aggregates the sequential reward, conditional reward, and the remaining constraint, resulting in $R=\mathrm{Avg}(R_s,R_c,r_7)=\mathrm{Avg}(0.33,1,0)=0.44$. 

This example shows that \ourmethod produces a faithful reward signal by penalizing violated prerequisites, selecting only active branches, and aggregating independent constraint groups according to the logic structure.

\section{Conclusion}

In this work, we propose \ourmethod, a training framework for logic-structured  instruction following. 
\ourmethod explicitly models logical relationships among constraints in both data construction and reward design, covering parallel, sequential, conditional, and nested structures. 
It further introduces structure-aware reward aggregation that follows the execution semantics of logic-structured constraints. Experimental results show that \ourmethod significantly improves instruction-following ability and logic reasoning ability. Interpretability analysis reveals that these gains are associated with changes in attention, with the model assigning more attention to constraint-related tokens and logical connectors, which also helps explain its improved performance on logic reasoning tasks.


\section*{Limitations}
Our study has the following main limitations. First, due to computational constraints, we do not evaluate our method on larger models such as 70B+, and validation at this scale would further strengthen the credibility and robustness of our approach. Second, our training data is primarily English. While results on CFBench suggest that our training can generalize to other languages, incorporating multilingual logic-structured instruction data would further strengthen the cross-lingual generalization claim.







\bibliography{ref}

\appendix
\clearpage
\newpage
\section*{Appendix}




\begin{algorithm*}[t]
\small
\caption{Logic-Structured Instruction Construction and Structure-Aware Reward Aggregation}
\label{alg:logic_structured_training}
\begin{algorithmic}[1]
\Require Seed instruction set $\mathcal{D}^{\mathrm{seed}}$, policy model $\pi_\theta$, reference policy $\pi_{\mathrm{ref}}$, group size $G$, sequential penalty factor $\lambda$, stability constant $\delta$
\Ensure Optimized policy model $\pi_\theta$

\Statex \textbf{Stage 1: Logic-structured instruction construction}

\For{each seed instruction $x^{\mathrm{seed}} \in \mathcal{D}^{\mathrm{seed}}$}
    \State Select task-compatible atomic constraints $\mathcal{C}_x$
    \State Select one or more logical structures from $\{\textsc{Par}, \textsc{Seq}, \textsc{Cond}, \textsc{Nest}\}$
    \State Compose $\mathcal{C}_x$ and the selected logical structures into a constraint tree $\mathcal{T}_x=(\mathcal{V},\mathcal{E},\rho)$
    \State Render $x^{\mathrm{seed}}$ together with $\mathcal{T}_x$ into a complex instruction $x$
    \State Construct the corresponding evaluator set $\mathcal{Q}_x$
    \State Store the training example $(x,\mathcal{T}_x,\mathcal{C}_x,\mathcal{Q}_x)$
\EndFor

\Statex \textbf{Stage 2: Structure-aware reward aggregation and policy optimization}

\For{each training instruction $x$}
    \State Sample a group of responses $\{y_i\}_{i=1}^{G}$, where $y_i \sim \pi_\theta(\cdot \mid x)$

    \For{$i=1$ to $G$}
        \For{each leaf node $\ell$ in $\mathcal{T}_x$}
            \State Evaluate the atomic constraint $c_\ell$ using its evaluator $q_\ell$
            \State Obtain the constraint-level reward $r_\ell = q_\ell(x,y_i,c_\ell)$, where $r_\ell \in \{0,1\}$
        \EndFor
        \State Compute the structured reward $R_i = R_{\mathrm{final}}(x,y_i)=\Call{Aggregate}{\rho}$
    \EndFor

    \State Compute the group mean $\mu_R$ and standard deviation $\sigma_R$ of $\{R_i\}_{i=1}^{G}$
    \For{$i=1$ to $G$}
        \State Compute the normalized advantage $\hat{A}_i=(R_i-\mu_R)/(\sigma_R+\delta)$
    \EndFor

    \State Update $\pi_\theta$ with the GRPO clipped objective using $\{\hat{A}_i\}_{i=1}^{G}$
    \State Regularize $\pi_\theta$ toward the reference policy $\pi_{\mathrm{ref}}$
\EndFor

\State \Return $\pi_\theta$

\Statex

\Function{Aggregate}{$v$}
    \If{$v$ is a leaf node}
        \State \Return $r_v$
    \EndIf

    \State Let $\mathrm{Ch}(v)=(u_1,\ldots,u_m)$ be the children of $v$

    \If{$\tau(v)=\textsc{Par}$}
        \State \Return the average child reward $\frac{1}{m}\sum_{j=1}^{m}\Call{Aggregate}{u_j}$

    \ElsIf{$\tau(v)=\textsc{Seq}$}
        \For{$j=1$ to $m$}
            \State Compute $a_j=\Call{Aggregate}{u_j}$ and $z_j=\mathbb{I}[a_j=1]$
            \State Compute the penalty factor $g_j=\lambda^{\sum_{k<j}(1-z_k)}$
        \EndFor
        \State \Return the order-aware reward $\frac{1}{m}\sum_{j=1}^{m} g_j a_j$

    \ElsIf{$\tau(v)=\textsc{Cond}$}
        \State Determine the activated branch $u_{v,\mathrm{act}}$ according to the branch condition
        \State \Return $\Call{Aggregate}{u_{v,\mathrm{act}}}$

    \ElsIf{$\tau(v)=\textsc{Nest}$}
        \State Recursively compute the reward of each child logical substructure
        \State \Return the aggregation result obtained by applying the top-level logical relation encoded in $v$
    \EndIf

\EndFunction

\end{algorithmic}
\end{algorithm*}

\section{Baselines}
\label{appx:bl}
\textbf{RAIF-7B}~\cite{qin2025incentivizing}: RAIF-7B (Incentivizing Reasoning) proposes a systematic approach to enhance large language models' ability to handle complex instructions by incentivizing reasoning processes during test-time computation scaling. The method encourages models to engage in explicit reasoning steps when processing complex instructions, thereby improving instruction-following performance through enhanced computational reasoning capabilities.

\textbf{Conifer-7B-DPO}~\cite{sun2024conifer}: Conifer addresses complex constrained instruction-following through a two-stage training pipeline. The method first constructs a curriculum dataset organized from simple to complex instructions and performs supervised fine-tuning (SFT) on this dataset. Subsequently, it applies Direct Preference Optimization (DPO) training using an open-source preference dataset to further refine the model's ability to follow complex constraints.

\textbf{Crab-7B-DPO}~\cite{qi2024constraint}: Crab employs a constraint back-translation strategy to improve complex instruction following. The method leverages Llama3-70B-Instruct as a strong teacher model to back-translate constraints into high-quality instruction-response pairs. This process creates a comprehensive dataset with complex constraints, which is then used for DPO training to enhance the model's instruction-following capabilities.

\textbf{SPAR-8B-DPO}~\cite{cheng2024spar}: SPAR (Self-play with tree-search refinement) introduces a self-play framework that integrates tree-search-based self-refinement mechanisms. The framework enables an LLM to play against itself, employing tree-search strategies to iteratively refine responses with respect to given instructions. This approach generates valid and comparable preference pairs while minimizing unnecessary variations, facilitating effective DPO training for instruction-following tasks.

\textbf{VERIF}~\cite{peng2025verif}: VERIF (Verification Engineering for Reinforcement Learning) combines multiple verification approaches to enhance instruction following through reinforcement learning. The method integrates rule-based code verification with LLM-based verification from large reasoning models (RLVR), providing comprehensive verification signals that guide the reinforcement learning process toward better instruction-following performance.

\textbf{Self-Supervised-7B}~\cite{ren2025instructions}: Self-Supervised-7B presents a self-supervised reinforcement learning framework for instruction following that eliminates the need for external supervision. The method extracts reward signals directly from instructions and generates pseudo-labels for reward model training, thereby removing dependencies on human-annotated preference data. The framework introduces constraint decomposition strategies and efficient constraint-level binary classification methods to address sparse reward problems while maintaining computational efficiency. Experimental results demonstrate significant performance improvements across multiple datasets, including complex agentic tasks and multi-turn instruction-following scenarios.

\section{Training Details}
\label{appx:tds}
\subsection{SFT Training}

We perform supervised fine-tuning (SFT) on six models: Qwen2.5-1.5B-Instruct, Qwen2.5-7B-Instruct, Llama-3.1-8B-Instruct, Distill-Qwen-7B, Distill-Qwen-14B, and Qwen3-8B. Training is conducted on 8 NVIDIA H200 GPUs. 

The training data consists of instruction-response pairs, where the responses are generated by teacher models. Specifically, GPT-4.1 is used as the teacher model for Qwen2.5-1.5B-Instruct, Qwen2.5-7B-Instruct, and Llama-3.1-8B-Instruct; QwQ-32B is used as the teacher model for Distill-Qwen-7B and Distill-Qwen-14B; and Qwen3-32B is used as the teacher model for Qwen3-8B. 

Training is conducted using LLaMA-Factory~\cite{zheng2024llamafactory} with LoRA fine-tuning (rank=8, targeting all linear layers). We use a maximum sequence length of 20480 tokens. Training hyperparameters include: batch size of 1 per device with gradient accumulation of 8 steps, learning rate of 1.0e-4, 3 training epochs and cosine learning rate scheduler with 10\% warmup ratio.

\subsection{RL Training}
\label{appx:rl}

\begin{table*}
\centering
\begin{minipage}{0.8\linewidth}
\resizebox{\linewidth}{!}{%
\begin{tabular}{l c}
\toprule
\textbf{Hyperparameter} & \textbf{Value} \\
\midrule
Training Framework & EasyR1~\cite{zheng2025easyr1} \\
\midrule
Global Batch Size & 96 \\
Micro Batch Size Per Device For Update & 4 \\
Micro Batch Size Per Device For Experience & 8 \\
Use KL Loss & True \\
Rollout Batch Size & 384 \\
Rollout n & 5 \\
Max Prompt Length & 2048 \\
Max Response Length & 8192 \\
\midrule
Training Epochs & \\
\quad Qwen2.5-7B-Instruct & 1 \\
\quad Llama-3.1-8B-Instruct & 1 \\
\quad Distill-Qwen-14B & 1 \\
\quad Distill-Qwen-7B & 5 \\
\quad Qwen2.5-1.5B-Instruct & 5 \\
\quad Qwen3-8B & 4 \\
\midrule
Actor Learning Rate & $1 \times 10^{-6}$ \\
KL Coefficient & $1 \times 10^{-2}$ \\
\midrule
Number of GPUs & 
\begin{tabular}[c]{@{}c@{}}
8 H200 for GRPO Training \\
1 H200 for Reward Model Flask Service
\end{tabular}
\\
\bottomrule
\end{tabular}%
}
\caption{GRPO training hyperparameters.}
\label{tab:aaaa}
\end{minipage}
\end{table*}

We implement GRPO training using the EasyR1 framework. All hyperparameters used in our experiments are summarized in Table~\ref{tab:aaaa}.

\subsection{Reward Model Training}
\label{appx:rmt}
We fine-tune Qwen2.5-7B-Instruct for a binary classification task to determine whether a response satisfies a given constraint, using the method and data from \citealp{ren2025instructions}. Training data consists of response-constraint pairs. Each sample is tokenized by concatenating the response and constraint into a single text sequence. We use full-parameter fine-tuning with the Hugging Face Trainer framework. Training hyperparameters: learning rate 5e-6, batch size 1 per device, gradient accumulation steps 1, 3 epochs, FP16 precision, gradient checkpointing enabled, and DeepSpeed optimization configured via JSON. The training is performed on 8 NVIDIA H200 GPUs.

We then construct 600 test samples, each consisting of an instruction with three atomic soft constraints, and use Llama-3.1-8B-Instruct to generate the corresponding response.
For each generated response, three human annotators label whether each atomic soft constraint is satisfied, resulting in 1,800 atomic constraint annotations in total.
Disagreements are resolved by majority voting.
The trained reward model predicts whether each atomic constraint is satisfied or unsatisfied.
We evaluate the reward model using Precision, Recall, and F1-score.
As shown in Tab.~\ref{tab:reward_model_evaluation}, the trained reward model achieves a Precision of 0.87, a Recall of 0.84, and an F1-score of 0.86, showing reasonable agreement with human judgment.
This indicates that the reward signals used for optimization are generally aligned with human evaluations.

\begin{table}[htbp]
\centering
\begin{tabular}{lc}
\toprule
\textbf{Metric} & \textbf{Value} \\
\midrule
Precision & 0.87 \\
Recall    & 0.84 \\
F1-score  & 0.86 \\
\bottomrule
\end{tabular}
\caption{Evaluation results of the trained reward model on atomic soft-constraint satisfaction prediction. Each instruction contains three atomic soft constraints.}
\label{tab:reward_model_evaluation}
\end{table}

\section{Benchmarks}
\label{appx:bec}

We evaluate instruction-following ability on the following benchmarks.

\textbf{IFEval}~\cite{zhou2023instruction} evaluates compliance with automatically verifiable instructions, covering 25 types of rule-based constraints across around 500 prompts.

\textbf{CFBench}~\cite{zhang2024cfbench} is a large-scale constraint-following benchmark with 1,000 samples spanning over 200 real-life scenarios and more than 50 NLP tasks. It categorizes constraints into 10 primary categories and over 25 subcategories.

\textbf{FollowBench}~\cite{jiang2023followbench} evaluates fine-grained instruction following across five constraint types: Content, Situation, Style, Format, and Example. It incrementally increases task difficulty by adding constraints level by level.

\textbf{ComplexBench}~\cite{wen2024benchmarking} assesses the ability of LLMs to follow complex instructions with multiple constraints. It provides a hierarchical taxonomy covering 4 constraint types, 19 constraint dimensions, and 4 composition types.

\textbf{Collie}~\cite{yao2023collie} evaluates constrained language generation through a grammar-based framework, supporting compositional constraints at word, sentence, paragraph, and passage levels.

\textbf{AgentIF}~\cite{qi2025agentif} evaluates instruction following in agentic scenarios. It contains long and complex instructions collected from 50 real-world agent applications, with an average of 11.9 constraints per instruction.

We further assess general reasoning and knowledge capabilities using the following benchmarks.

\textbf{GPQA-Diamond}~\cite{rein2024gpqa} is a challenging subset of GPQA, containing 198 graduate-level multiple-choice questions in biology, chemistry, and physics.

\textbf{AIME2024} and \textbf{AIME2025} consist of problems from the corresponding American Invitational Mathematics Examination competitions and are used to evaluate mathematical reasoning.

\textbf{Enigmata}~\cite{chen2025enigmata} is a logical reasoning benchmark with 36 tasks across seven categories. Each task includes automatic generators and rule-based verifiers, enabling scalable and fine-grained evaluation.

\textbf{MT-Bench}~\cite{zheng2023judging} evaluates multi-turn conversational ability using 80 open-ended questions covering diverse topics such as writing, role-playing, mathematics, and coding.

\textbf{AlpacaEval 2.0}~\cite{dubois2024length} is an automated instruction-following benchmark that uses GPT-4-based evaluation to compare model responses with reference outputs.

\section{Dataset}
\label{appx:ctd}


\begin{table}[t]
\centering

\setlength{\tabcolsep}{6pt}
\renewcommand{\arraystretch}{1.05}
\begin{tabular}{|p{\dimexpr\columnwidth-2\tabcolsep-2\arrayrulewidth\relax}|}

\hline

\textbf{[Task Description]}
\begin{enumerate}[leftmargin=1.5em, itemsep=0pt, topsep=2pt]
    \item I currently have a seed question, but the seed questions are relatively simple. 
    To make the instructions for the seed question more complex, I want you to add more 
    constraints to this question.
    
    \item I will provide [Seed Question] and [Constraint References], and you can use these 
    constraint references to increase the difficulty of the seed question.
    
    \item \textbf{[Constraint References]} are just suggestions for constraints. When adding 
    constraints, you can add one or more, freely combine from the constraint references, 
    or add other constraints you deem appropriate.
    
    \item Do not delete any information from the seed question. Your task is to rewrite the 
    seed question and add constraints without omitting any key information from the seed question, 
    such as reference texts.
    
    \item Directly return the modified question, the question with added constraints, without 
    any analysis.
\end{enumerate}
\tabularnewline

\textbf{[Constraint References]} \newline
1. Lexical content constraint : \{Definition\} \{Example\} \newline
2. Morphological constraint : \{Definition\} \{Example\} \newline
... ...
\tabularnewline

\textbf{[Seed Question] :} \texttt{\{raw\_question\}}
\tabularnewline
\hline

\textbf{[Modified Question] :}
\tabularnewline
\hline

\end{tabular}
\caption{Prompt template for adding atomic constraints.}
\label{tab:atomic-template}
\end{table}

As shown in Tab.~\ref{tab:ht} and Tab.~\ref{tab:st}, we distinguish between \textit{soft} and \textit{hard} constraints on LLM outputs. Soft constraints cannot be reliably verified by fixed symbolic rules, as they target high-level, often subjective properties such as semantic focus, tone and emotion, stylistic form, audience- or author-specific style, and syntactic patterns. In contrast, hard constraints are explicitly rule-checkable: they specify concrete requirements on keywords and their frequencies, lengths (in words, sentences, or paragraphs), detectable formats (e.g., numbered bullets, titles, JSON), presence of placeholders or postscripts, and strict start/end markers or punctuation usage. We provide the prompt template for adding atomic constraints in Tab.~\ref{tab:atomic-template}. 

\begin{table*}[t]
\centering
\footnotesize
\begin{tabular}{llp{7cm}}
\toprule
\textbf{Instruction Group} & \textbf{Instruction} & \textbf{Description} \\
\midrule
\multirow{5}{*}{Keywords} 
& Include Keywords & Response must include specified keywords (e.g., \texttt{\{keyword1\}, \{keyword2\}}). \\
& Keyword Frequency & A particular word should appear a certain number of times (\texttt{\{N\} times}). \\
& Forbidden Words & Prohibits the inclusion of specified keywords (\texttt{\{forbidden words\}}). \\
& Letter Frequency & Requires a specific letter to appear a certain number of times (\texttt{\{N\} times}). \\
& Response Language & Entire response must be in a specified language (\texttt{\{language\}}) and no other. \\
\cmidrule{1-3}
\multirow{4}{*}{Length Constraints}
& Number Paragraphs & Specifies the exact number of paragraphs (\texttt{\{N\}}), separated by markdown divider \texttt{***}. \\
& Number Words & Constraint on the number of words: ``at least / around / at most \texttt{\{N\} words}''. \\
& Number Sentences & Constraint on the number of sentences: ``at least / around / at most \texttt{\{N\} sentences}''. \\
& Number Paragraphs + First Word & Requires \texttt{\{N\}} paragraphs (separated by two line breaks), with the \texttt{\{i\}}-th paragraph starting with a specified word (\texttt{\{first\_word\}}). \\
\cmidrule{1-3}
\multirow{2}{*}{Detectable Content}
& Postscript & Requires an explicit postscript at the end, starting with a specified marker (\texttt{\{postscript marker\}}). \\
& Number Placeholder & Response must contain at least \texttt{\{N\}} placeholders in square brackets (e.g., \texttt{[address]}). \\
\cmidrule{1-3}
\multirow{6}{*}{Detectable Format}
& Number Bullets & Requires exactly \texttt{\{N\}} bullet points using markdown format (e.g., \texttt{* This is a point.}). \\
& Title & Answer must include a title wrapped in double angular brackets (e.g., \texttt{<<poem of joy>>}). \\
& Choose From & Response must be one of the provided options (\texttt{\{options\}}). \\
& Minimum Number Highlighted Section & Requires at least \texttt{\{N\}} sections highlighted using markdown (e.g., \texttt{*highlighted section*}). \\
& Multiple Sections & Response must have \texttt{\{N\}} sections, with each section's beginning marked by a splitter (e.g., \texttt{\{section\_splitter\} X}). \\
& JSON Format & Entire output must be wrapped in JSON format. \\
\cmidrule{1-3}
\multirow{2}{*}{Combination}
& Repeat Prompt & First repeat the request without change, then provide the answer. \\
& Two Responses & Requires two different responses, separated by six asterisk symbols (\texttt{******}). \\
\cmidrule{1-3}
\multirow{3}{*}{Change Cases}
& All Uppercase & Entire response must be in English, using only capital letters. \\
& All Lowercase & Entire response must be in English, using only lowercase letters, with no capital letters allowed. \\
& Frequency of All-capital Words & Words with all capital letters should appear ``at least / around / at most \texttt{\{N\} times}''. \\
\cmidrule{1-3}
\multirow{2}{*}{Start with / End with}
& End Checker & Response must end with a specific phrase (\texttt{\{end\_phrase\}}), with no other words following it. \\
& Quotation & Entire response must be wrapped in double quotation marks. \\
\cmidrule{1-3}
Punctuation & No Commas & Prohibits the use of any commas in the entire response. \\
\bottomrule
\end{tabular}
\caption{Hard Constraint Types.}
\label{tab:ht}
\end{table*}

\begin{table*}
\centering
\renewcommand{\arraystretch}{1.5}
\small
\begin{tabular}{p{3.5cm}|p{4.8cm}|p{5.5cm}}
\hline
\textbf{Constraint Type} & \textbf{Definition} & \textbf{Example} \\
\hline
Element constraint & Requires inclusion of specific entities or scenarios. & ``...highlights the Great Wall.'' \\
\hline
Semantic constraint & Focuses on themes, tone, or stance. & ``Write a poem about London.'' \\
\hline
Paragraph Count & Limits the number of paragraphs. & ``...divided into 3 sections.'' \\
\hline
Document Count & Limits the number of documents. & ``...list 3 articles.'' \\
\hline
Tone and emotion & Conforms to specific emotional tone. & ``Write a letter in an angry and sarcastic tone.'' \\
\hline
Form and style & Uses specified stylistic form and perception. & ``Write a passage in an encyclopedic style.'' \\
\hline
Audience-specific & Tailored to a specific audience group. & ``Write a poem for a 6-year-old.'' \\
\hline
Authorial style & Emulates specific authors' styles. & ``Write a passage in the style of Shakespeare.'' \\
\hline
Bespoke format & Uses custom formatting protocols. & ``Bold the main idea and output in unordered list.'' \\
\hline
Specialized format & Tailored for specific applications or domains. & ``Convert to electronic medical record format.'' \\
\hline
Pragmatic constraint & Adapts to context like dialects or language policy. & ``Output in English, classical Chinese, etc.'' \\
\hline
Syntactic constraint & Follows specific phrase and clause structures. & ``Use imperatives with nouns and verb phrases.'' \\
\hline
Morphological constraint & Controls affixes, roots, and word formation. & ``Output all content in lowercase English.'' \\
\hline
Phonological constraint & Focuses on sounds, tone, and intonation. & ``Single-syllable tongue twisters.'' \\
\hline
Role-based constraint & Responds with specific role identity. & ``You are Confucius, how do you decide?'' \\
\hline
Task-specific constraint & Addresses a defined situational task. & ``Work from home, how to report?'' \\
\hline
Complex context constraint & Involves multi-faceted and nested reasoning. & ``On the left, 10 total, what to do?'' \\
\hline
Example constraint & Conforms to patterns from example pairs. & ``input:x..., output:\{...\}; input:y..., output?'' \\
\hline
Inverse constraint & Narrows response space via exclusions. & ``No responses about political topics.'' \\
\hline
Contradictory constraint & Combines requirements that are hard to satisfy simultaneously. & ``A five-character quotation, 1000 words.'' \\
\hline
Rule constraint & Follows symbolic or logical operation rules. & ``Each answer adds 1+1=3, then 2+2=5.'' \\
\hline
\end{tabular}
\caption{Soft Constraint Types.}
\label{tab:st}
\end{table*}

\section{Sensitivity Analysis on Penalty Factor}
\label{appx:sen}

We conduct a sensitivity study on the penalty factor $\lambda$ using Qwen2.5-7B-Instruct.
As shown in Tab.~\ref{tab:decay_sensitivity}, our setting $\lambda=0.5$ provides appropriate penalty and achieves the best performance.

\begin{table}[htbp]
\centering
\begin{tabular}{lcc}
\toprule
\textbf{Setting} & \textbf{IFEval} & \textbf{CFBench} \\
\midrule
Base      & 73.9 & 47.0 \\
$\lambda = 0.3$ & 78.4 & 52.0 \\
\textbf{$\lambda = 0.5$} & \textbf{79.7} & \textbf{54.0} \\
$\lambda = 0.7$ & 77.4 & 51.0 \\
$\lambda = 1$   & 75.0 & 49.0 \\
\bottomrule
\end{tabular}
\caption{Sensitivity analysis of the decay coefficient $\lambda$.}
\label{tab:decay_sensitivity}
\end{table}

\section{Performance on Nested Logical Structures}

We evaluate model performance on the nested logical-structure test set of ComplexBench, which includes \texttt{Selection\_and\_Chain\_2} and \texttt{Selection\_and\_Chain\_3}. These two subsets denote nested structures with depths of 2 and 3, respectively. As shown in Tab.~\ref{tab:nested_complexbench}, our method consistently improves performance on nested structures.
Notably, after training, Distill-Qwen-14B achieves a 14.4\% improvement on \texttt{Selection\_and\_Chain\_3}.

\begin{table}[htbp]
\centering
\resizebox{\linewidth}{!}{
\begin{tabular}{lcc}
\toprule
\textbf{Model} & \textbf{\texttt{Select\_and\_Chain\_2}} & \textbf{\texttt{Select\_and\_Chain\_3}} \\
\midrule
Distill-Qwen-7B        & 42.7 & 40.0 \\
Distill-Qwen-7B-\ourmethod  & \textbf{44.3} \textcolor{red}{(+1.6)} & \textbf{45.9} \textcolor{red}{(+5.9)} \\
\midrule
Distill-Qwen-14B       & 64.1 & 54.7 \\
Distill-Qwen-14B-\ourmethod & \textbf{68.7} \textcolor{red}{(+4.6)} & \textbf{69.1} \textcolor{red}{(+14.4)} \\
\bottomrule
\end{tabular}
}
\caption{Performance on the nested logical-structure test set of ComplexBench.}
\label{tab:nested_complexbench}
\end{table}

\begin{figure*}[t] 
    \centering
    \begin{subfigure}[b]{0.35\textwidth}
        \centering
        \includegraphics[width=\textwidth]{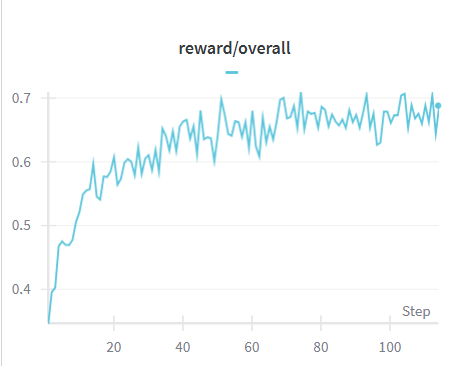}
        \caption{Reward Dynamics of Qwen2.5-7B-Instruct.}
        \label{fig:before_inst}
    \end{subfigure}
    \hfill
    \begin{subfigure}[b]{0.35\textwidth}
        \centering
        \includegraphics[width=\textwidth]{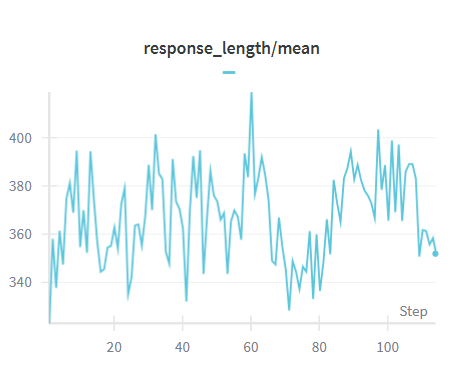}
        \caption{Response Length Dynamics of Qwen2.5-7B-Instruct.}
        \label{fig:after_inst}
    \end{subfigure}
    
    
    \begin{subfigure}[b]{0.35\textwidth}
        \centering
        \includegraphics[width=\textwidth]{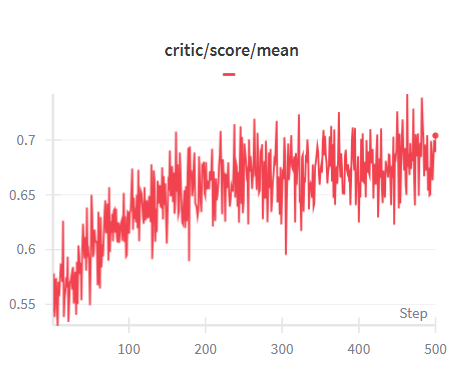}
        \caption{Reward Dynamics of Distill-Qwen-7B.}
        \label{fig:before_logic}
    \end{subfigure}
    \hfill
    \begin{subfigure}[b]{0.35\textwidth}
        \centering
        \includegraphics[width=\textwidth]{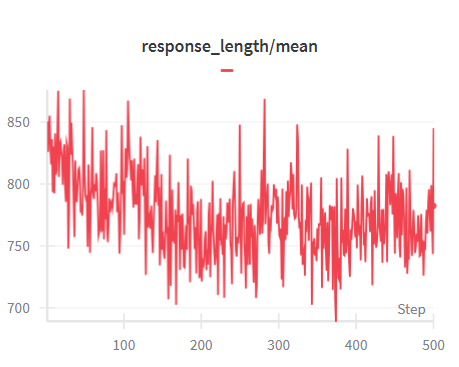}
        \caption{Response Length Dynamics of Distill-Qwen-7B.}
        \label{fig:after_logic}
    \end{subfigure}

    \begin{subfigure}[b]{0.35\textwidth}
        \centering
        \includegraphics[width=\textwidth]{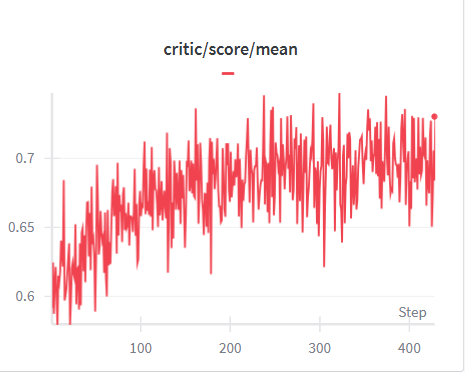}
        \caption{Reward Dynamics of Distill-Qwen-14B.}
        \label{fig:before_inst}
    \end{subfigure}
    \hfill
    \begin{subfigure}[b]{0.35\textwidth}
        \centering
        \includegraphics[width=\textwidth]{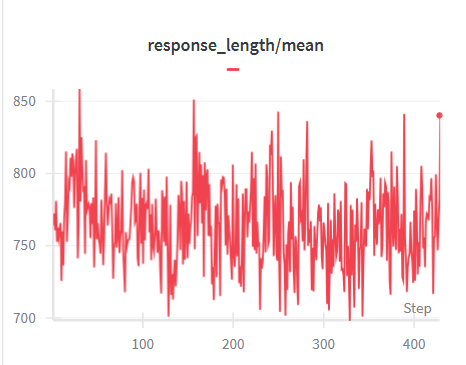}
        \caption{Response Length Dynamics of Distill-Qwen-14B.}
        \label{fig:after_inst}
    \end{subfigure}
    
    
    \begin{subfigure}[b]{0.35\textwidth}
        \centering
        \includegraphics[width=\textwidth]{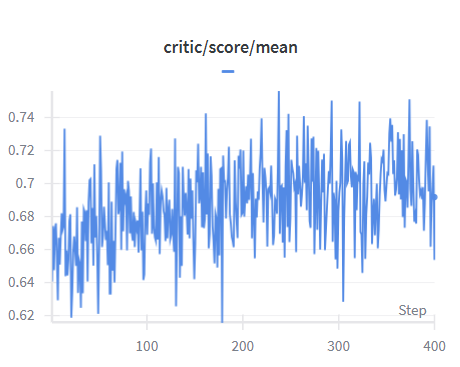}
        \caption{Reward Dynamics of Qwen3-8B.}
        \label{fig:before_logic}
    \end{subfigure}
    \hfill
    \begin{subfigure}[b]{0.35\textwidth}
        \centering
        \includegraphics[width=\textwidth]{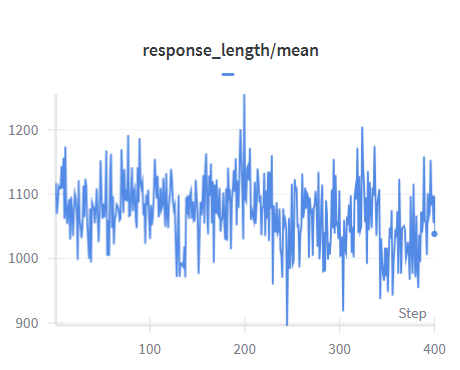}
        \caption{Response Length Dynamics of Qwen3-8B.}
        \label{fig:after_logic}
    \end{subfigure}
    
    \caption{Training dynamics of reward and response length.}
    \label{fig:case13}
\end{figure*}

\begin{figure}[!t]
    \centering
    \begin{subfigure}[!t]{\linewidth}
        \centering
        \includegraphics[width=\linewidth]{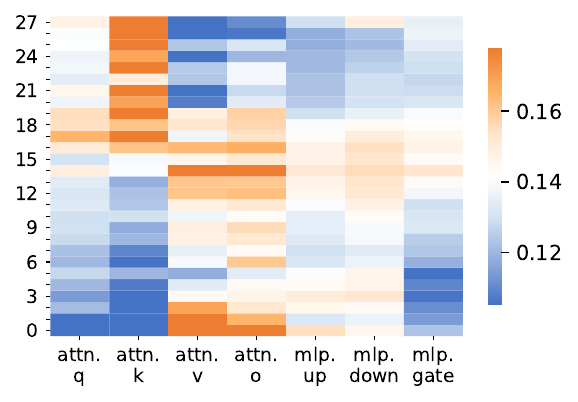}
        \caption{Qwen2.5-1.5B-Instruct}
        \label{fig:param-qwen-7b1}
    \end{subfigure}
    \quad
    \begin{subfigure}[!t]{\linewidth}
        \centering
        \includegraphics[width=\linewidth]{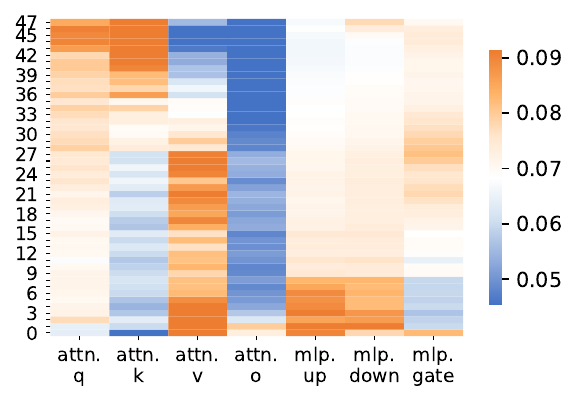}
        \caption{Distill-Qwen-14B}
        \label{fig:param-distill-7b1}
    \end{subfigure}
    \quad
    \begin{subfigure}[!t]{\linewidth}
        \centering
        \includegraphics[width=\linewidth]{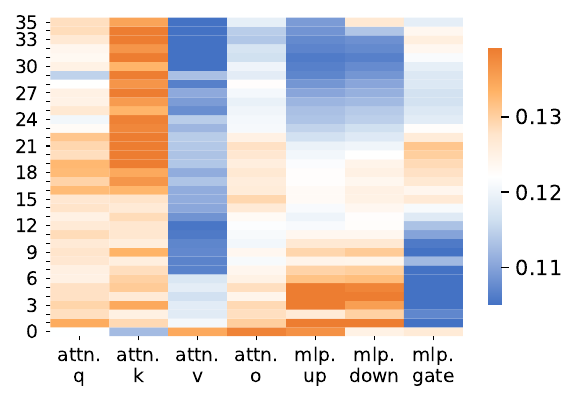}
        \caption{Qwen3-8B}
        \label{fig:param-distill-7b1}
    \end{subfigure}

    \caption{Parameter change rates of LLMs to the original ones across different modules.
    }
    \label{fig:parameter_app}
\end{figure}

\section{Training Dynamics Analysis}

Fig.~\ref{fig:case13} shows the reward and response-length dynamics during \ourmethod training. Across all models, rewards exhibit a clear upward trend in the early stage and then gradually stabilize with moderate oscillations. Qwen2.5-7B-Instruct improves rapidly within the first few steps, while Distill-Qwen-7B and Distill-Qwen-14B show steadier gains over longer training. Qwen3-8B also maintains an overall increasing trend, despite larger fluctuations.

In contrast, response lengths remain highly variable and do not show a consistent increasing pattern. Qwen2.5-7B-Instruct produces relatively short responses, while Distill-Qwen and Qwen3-8B generate longer outputs. Notably, the reward improvements are not accompanied by monotonic length growth, suggesting that \ourmethod improves constraint satisfaction rather than simply encouraging longer responses.

\section{Full Parameter Change Patterns} \label{appx:full}

As shown in Fig.~\ref{fig:parameter_app}, we extend the parameter change analysis to Qwen2.5-1.5B-Instruct, Distill-Qwen-14B, and Qwen3-8B. Across models, parameter updates exhibit clear module- and layer-dependent patterns rather than uniform changes. In the attention modules, query and key projections often show more pronounced changes, especially in middle-to-upper layers. This suggests that \ourmethod tends to affect components related to token interaction and attention pattern formation.

At the same time, value and output projections show model-specific behaviors. For example, Distill-Qwen-14B exhibits noticeable changes in the value projection across several lower and middle layers, while the output projection remains comparatively stable in many layers. Qwen2.5-1.5B-Instruct and Qwen3-8B also show localized changes in value/output projections, but the affected layer ranges differ across models.

MLP modules present smoother and more localized changes compared with attention modules. In particular, MLP up/down/gate projections tend to change more in certain lower or middle layers, depending on the model. Overall, these results indicate that \ourmethod induces structured parameter adaptations across both attention and MLP modules, with attention-related projections showing stronger layer-wise heterogeneity.

\begin{figure*}[t] 
    \centering
    
    
    \begin{subfigure}[b]{0.8\textwidth}
        \centering
        \includegraphics[width=\textwidth]{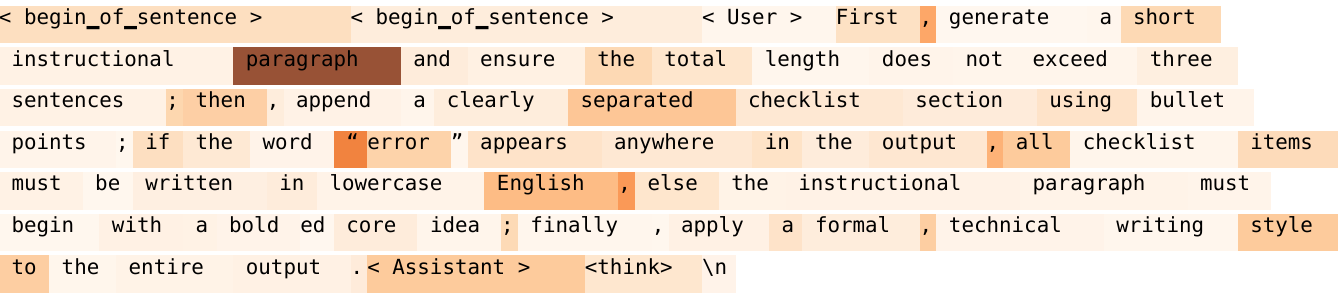}
        \caption{Before Training - Distill-Qwen-7B}
        \label{fig:before_inst}
    \end{subfigure}
    \hfill
    \begin{subfigure}[b]{0.8\textwidth}
        \centering
        \includegraphics[width=\textwidth]{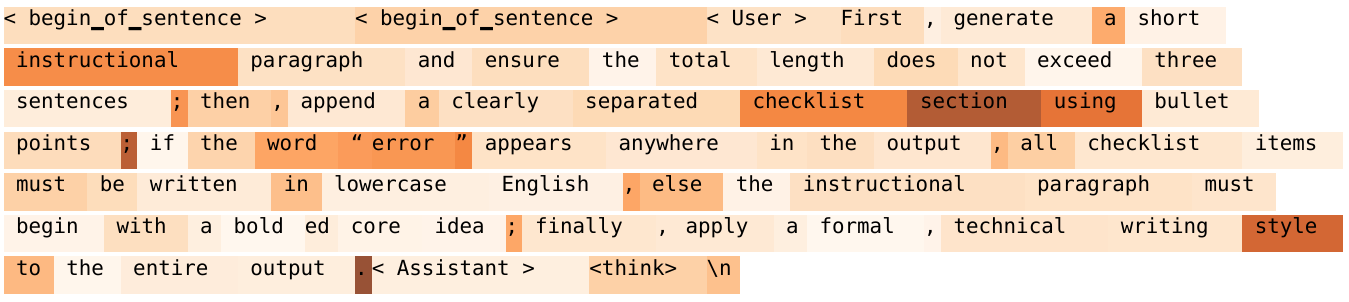}
        \caption{After Training - Distill-Qwen-7B}
        \label{fig:after_inst}
    \end{subfigure}
    
    
    \begin{subfigure}[b]{0.8\textwidth}
        \centering
        \includegraphics[width=\textwidth]{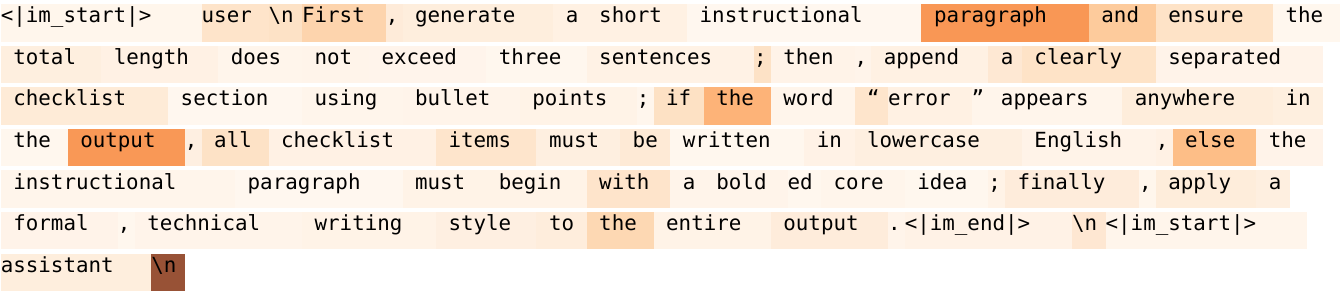}
        \caption{Before Training - Qwen3-8B}
        \label{fig:before_logic}
    \end{subfigure}
    \hfill
    \begin{subfigure}[b]{0.8\textwidth}
        \centering
        \includegraphics[width=\textwidth]{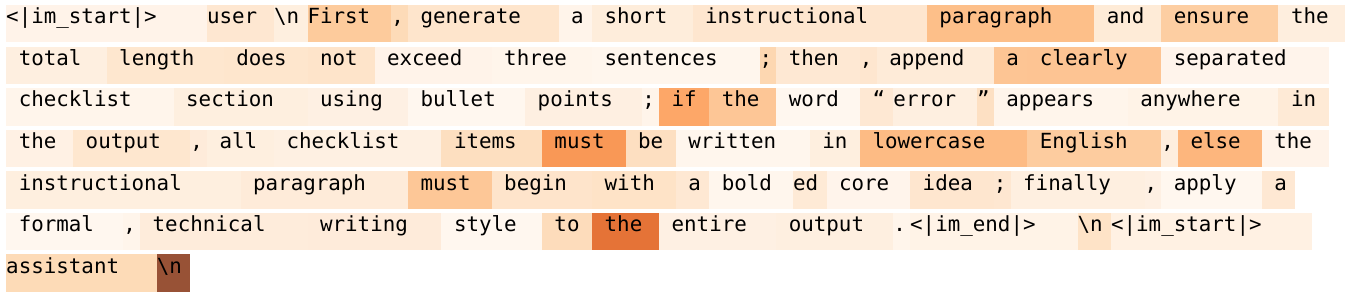}
        \caption{After Training - Qwen3-8B}
        \label{fig:after_logic}
    \end{subfigure}
    
    \caption{Token-level information flow analysis. Darker orange indicates higher gradient-based token saliency.
}
    \label{fig:case116}
\end{figure*}

\section{Full Token-Level Information Flow Analysis} \label{appx:full1}

As shown in Fig.~\ref{fig:case116}, we further analyze token-level information flow on Distill-Qwen-7B and Qwen3-8B. Before training, both models show relatively diffuse token-importance patterns, where salient tokens are distributed across several content and constraint-related words, such as ``instructional'', ``paragraph'', ``checklist'', ``error'', and ``output''. 

After training, token importance becomes more concentrated on instruction-critical elements. For Distill-Qwen-7B, stronger importance appears around ordering and conditional cues such as ``First'', ``then'', ``if'', and ``else'', as well as constraint-related words such as ``bullet'', ``lowercase'', ``final'', and ``style''. For Qwen3-8B, the after-training pattern similarly highlights key constraint and logic-related tokens, including ``First'', ``then'', ``if'', ``lowercase'', ``else'', ``formal'', and ``style''.

These qualitative examples suggest that \ourmethod encourages models to assign higher importance to tokens that define logical relations, ordering requirements, and formatting constraints, rather than treating the prompt uniformly. The trend is observed in both Distill-Qwen-7B and Qwen3-8B, indicating that logic-structured training may help models better identify and track key constraints in complex instructions.

\section{Ethical Considerations}

We discuss potential ethical concerns as follows. The annotation of the sampled soft-constraint dataset was conducted by three annotators recruited through our institution. The annotators were asked to judge whether each generated response satisfied the corresponding soft constraint, and disagreements were resolved by majority voting. To protect annotator privacy, their identities were anonymized from the authors, and no personally identifiable information was used in our analysis. All annotators were informed of the research purpose of the annotation task, consented to the use of the annotated data for research, and were compensated above the local minimum wage. The instructions given to annotators are shown in Tab.~\ref{tab:annotation_instruction}.
\begin{table}[t]
\centering
\small
\renewcommand{\arraystretch}{1.2}
\begin{tabular}{p{0.22\linewidth} p{0.70\linewidth}}
\toprule
\textbf{Section} & \textbf{Content} \\
\midrule
Annotation Task &
You will be shown an instruction, a soft constraint extracted from the instruction, and a model response. Your task is to judge whether the model response satisfies the given soft constraint. \\

\midrule
Please label each example as &
Satisfied: The response follows the soft constraint. The response does not need to be perfect, but the requirement should be clearly fulfilled. \\

\midrule
Please label each example as &
Unsatisfied: The response fails to follow the soft constraint, only partially follows it, or contradicts it. \\

\midrule
Additional instruction &
Please focus only on the provided soft constraint. Do not judge other aspects of the response, such as general helpfulness, factual correctness, or grammar, unless they are directly relevant to the soft constraint. For constraints about style, tone, sentiment, or level of detail, consider the overall response rather than a single word or phrase. If the case is ambiguous, choose the label that best reflects whether the response satisfies the constraint as a whole. \\
\bottomrule
\end{tabular}
\caption{Annotation instructions for soft-constraint satisfaction judgment.}
\label{tab:annotation_instruction}
\end{table}

\section{Declaration of LLM usage}
We acknowledge that \href{https://github.com/cursor/cursor}{Cursor} was used as an AI-assisted writing tool to refine the language of an early draft of this manuscript. All core ideas and contributions presented in this paper were independently developed by the authors.

\end{document}